\documentclass{article} 
\usepackage{PRIMEarxiv}

\usepackage[utf8]{inputenc} 
\usepackage[T1]{fontenc}    
\usepackage{hyperref}       
\usepackage{url}            
\usepackage{booktabs}       
\usepackage{amsfonts}       
\usepackage{nicefrac}       
\usepackage{microtype}      
\usepackage{amsmath}
\usepackage{amssymb}
\usepackage{xcolor}
\usepackage{fancyhdr}       
\usepackage{graphicx}       
\graphicspath{{media/}}     
\usepackage{wrapfig}
\usepackage{multirow, makecell}
\usepackage{natbib}
\usepackage{comment}

\title{FTD-GNO: Memory-Efficient Graph Neural Operators through Functional Tensor Decomposition \\ of the Kernel}

\author{
  Xiaomin~Zhang\textsuperscript{1},
  Boyue~Wang\textsuperscript{1},
  Junbin~Gao\textsuperscript{2}, 
  Yongli~Hu\textsuperscript{1},
  Baocai~Yin\textsuperscript{1} \\[1ex]
  \textsuperscript{1}Beijing Key Laboratory of Multimedia and Intelligent Software Technology \\
  Beijing Artificial Intelligence Institute \\
  School of Information Science and Technology \\
  Beijing University of Technology, Beijing 100124, China \\
  \texttt{zhxm\_6758@163.com},
  \texttt{\{wby,huyongli,ybc\}@bjut.edu.cn} \\[1ex]
  \textsuperscript{2}Discipline of Business Analytics \\
  The University of Sydney Business School \\
  The University of Sydney, Camperdown NSW 2006, Australia \\
  \texttt{junbin.gao@sydney.edu.au}
}

\begin{document}
\maketitle

\begin{abstract}
Graph Neural Operators (GNOs) provide flexible surrogate models for learning solution operators of partial differential equations (PDEs). 
However, standard GNOs typically parameterize the integral kernel with a monolithic neural network and evaluate kernel interactions over graph edges, leading to substantial computational and memory overhead at high resolutions or with large neighborhoods. 
To address these limitations, we propose Functional Tensor Decomposition Graph Neural Operator (FTD-GNO), a memory-efficient GNO framework that decouples the high-dimensional continuous integral kernel into low-dimensional mode-wise functions. By instantiating the kernel with classical tensor decomposition formats, including CP, Tensor-Train, and Tucker decompositions, FTD-GNO enables algebraic reconstruction of the integral operator without explicitly materializing full edge-wise kernel tensors. This factorized formulation reduces the memory footprint of kernel evaluation and aggregation while retaining the continuous operator-learning structure of GNOs. Theoretical complexity analysis shows that FTD-GNO substantially lowers parameter and activation-memory costs associated with high-dimensional kernel construction. Experiments show lower peak memory than the corresponding unfactorized graph-integral baselines, with shorter recorded training times. Fourier–graph experiments further demonstrate that FTD can improve the efficiency of a graph-integral layer within a hybrid operator and has good scalability.
\end{abstract}

\section{Introduction}\label{sec:intro}
Partial differential equations (PDEs) are the core mathematical tools for characterizing the spatiotemporal evolution of complex physical systems \citep{PDE1, PDE2}. 
In practical scientific and engineering applications, such as uncertainty quantification, optimal control, and inverse problem solving, it is often necessary to perform massive, repeated evaluations of PDEs driven by varying physical parameters. 
Traditional numerical solvers (e.g., finite element \citep{FEM1, FEM2} or finite difference methods \citep{FDM1, FDM2} rely heavily on high-resolution grids, and their prohibitive iterative computational overhead struggles to meet the demands of real-time inference in complex systems \citep{FDM-FEM1, FDM-FEM2}. 
Recently, operator learning has emerged as a novel deep learning paradigm. 
Unlike conventional neural networks that only learn mappings between finite-dimensional vectors, neural operators (NOs) \citep{NOs, NeuralOperator} aim to directly construct nonlinear mappings between infinite-dimensional continuous function spaces. 
This discretization-independent architecture endows models with the remarkable capability of zero-shot cross-resolution generalization. 

\begin{figure}
    \centering
    \includegraphics[width=0.65\linewidth]{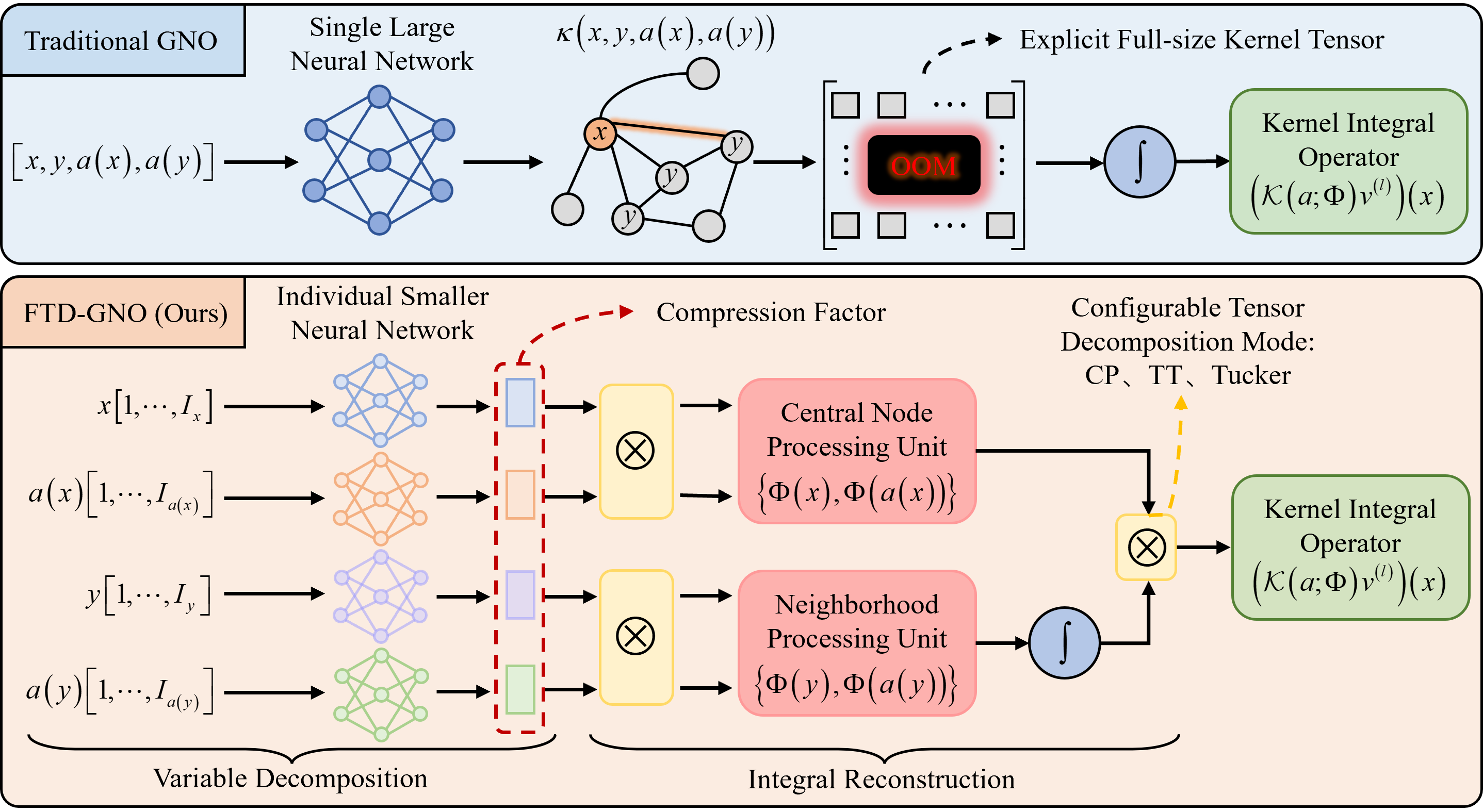}
    \caption{Efficient GNO via Function Tensor Decomposition: GNO calculates the integral kernel through a single large neural network and explicitly constructs the complete integral kernel tensor on each edge. The proposed methodology involves the decomposition of the integral kernel into multiple single-variable neural networks, in conjunction with the utilisation of configurable tensor decomposition modes. This approach facilitates the reconstruction of the integral operator, thereby circumventing the occurrence of memory overflow. }
    \label{fig1}
\end{figure}

Graph Neural Operators (GNOs) \citep{GNOs} discretize continuous spaces into graph topologies and approximate integral transforms via message-passing mechanisms, demonstrating unparalleled flexibility in handling unstructured grids and complex geometric domains. 
However, when confronted with high-dimensional systems and high-resolution grids, the underlying architecture of standard GNOs exposes intertwined flaws. 
Specifically, during the integral kernel evaluation and spatial aggregation phases, information interaction relies heavily on edge-based message-passing mechanisms. 
This implies that the grid must explicitly construct, instantiate and store the complete integral kernel tensor on each edge within the computational graph. 
As spatial dimensionality increases or grid nodes become denser, the memory requirement to construct the full input exhibits explosive exponential growth, manifesting as the ``curse of dimensionality". 
Such edge-level storage dependency severely limits network depth and renders GNOs highly susceptible to Out-Of-Memory (OOM) \citep{OOM} errors, profoundly restricting their practical utility. 
Furthermore, to adequately capture the complex feature couplings in multi-dimensional spaces, the hidden layer dimensions of the network are forced to expand drastically, triggering a nonlinear surge in model parameters. 

To address the aforementioned challenges, this paper proposes a novel Graph Neural Operator framework based on Functional Tensor Decomposition (FTD-GNO). 
The essence of this framework lies in the deep decoupling of the high-dimensional tensor construction process within the continuous domain. 
The overall schematic of this setup is illustrated in Figure \ref{fig1}. 
Specifically, we elegantly strip the originally massive full-scale network inputs into low-dimensional variable sets corresponding to independent physical dimensions. 
Under the multi-branch architecture of FTD-GNO, lightweight subnetworks tailored for specific dimensions only need to focus on the 1D feature objects within their respective dimensions. 
Additionally, we leverage the associativity of tensor decomposition to perform an algebraic reconstruction of the spatial integral operators, with source-dependent factors aggregated before their combination with query-dependent factors. 

This decomposition and reconstruction strategy thoroughly circumvents the explicit computation of full-size tensors. 
Without sacrificing the continuous spatial expressive capacity of the neural operator, it successfully compresses the memory complexity during the kernel evaluation and spatial integral aggregation stages from exponential to linear growth. 
Concurrently, benefiting from the low-rank factorized decoupling of multi-dimensional variables by the integral kernel, this architecture successfully addresses the bottleneck of exponential parameter explosion with respect to dimensionality. 
This has resulted in a substantial streamlining of the parameter space of the model. 

Under the proposed framework, we focus on combining Graph Neural Operators with three prominent tensor decomposition techniques, including \emph{Canonic-Polyadic (CP)} \citep{CP}, \emph{Tensor-Train (TT)} \citep{TT} and \emph{Tucker} decomposition \citep{Tucker}. 
The main contributions of our work are summarized as follows: 
\begin{itemize}
    \item \textbf{Parameter Compression via Tensor Decoupling:} 
    We integrate functional tensor decomposition into GNOs to decouple high-dimensional kernels into low-dimensional variables, drastically slashing the parameter scale. 
    \item \textbf{Memory Optimization via Algebraic Reconstruction:} 
    We algebraically reconstruct the integral operator to shift from edge-based message passing to node-based computation, greatly reducing memory costs. 
    \item \textbf{Comprehensive Empirical Validation:} 
    We demonstrate the highly competitive performance of the proposed method compared to GNOs in PDE benchmarks providing compelling evidence that firmly corroborates our theoretical analysis. 
\end{itemize}

\section{Related works}
\label{Sec2}

\textbf{Neural Operators }
Distinguished from traditional surrogate models for PDEs, the core advantage of neural operators lies in their construction of a universal approximator defined between function spaces. 
According to their architectures, they can be broadly categorized into three families \citep{NOzongshu}. 
The first family comprises operator architectures based on the universal approximation theorem. 
DeepONet \citep{DeepONet} pioneered a dual-path network architecture, encoding discretized sensor data of input functions via a branch network and processing target query coordinates via a trunk network, thereby laying the foundational theory for operator learning. 
Subsequently, architectures based on integral kernel transforms further expanded the boundaries of this field. 
Among them, Fourier Neural Operator (FNO) \citep{FNO} innovatively maps the integral kernel into the Fourier frequency domain, leveraging the Fast Fourier Transform (FFT) to extract low-frequency modes, thus achieving highly computationally efficient global convolutions. 
Meanwhile, Graph Neural Operator (GNO) \citep{GNOs} employs graph neural networks to perform Monte-Carlo integral approximations within the spatial domain, effectively modeling nonlocal operators on complex geometric domains. 
In recent years, models represented by GK-Transformer \citep{GK-Transformer} and General Neural Operator (GNOT) \citep{GNOT} have further integrated attention mechanisms into operator architectures, significantly enhancing the capability to capture features in multi-scale physical fields.

\textbf{Functional Tensor Decomposition in Deep Learning } 
Classical tensor decomposition methods have long been widely utilized to compress high-dimensional data matrices and reduce model ranks \citep{low-rank, tensor-decomposition}, including CP \citep{CP}, TT \citep{TT}, and Tucker \citep{Tucker} decomposition. 
In recent years, these algebraic tools have been introduced into deep learning models within the continuous function domain, spawning the emerging research direction of functional tensor decomposition (FTD). 
In the field of Implicit Neural Representations (INRs), frameworks such as F-INR \citep{F-INR} dismantle the massive Multi-Layer Perceptrons (MLPs) used for processing high-dimensional coordinates into a series of axially independent, lightweight subnetworks. 
This greatly alleviates the curse of dimensionality encountered when fitting complex videos or high-resolution scenes, and significantly accelerates convergence. 
Similarly, in Physics-Informed Neural Networks (PINNs) \citep{PINN} for solving PDEs, employing tensor decomposition formats to separate spatial and temporal variables not only improves the network's ability to fit gradients within complex loss landscapes but also effectively circumvents the computational dissipation caused by high-dimensional dense sampling. 
These advancements thoroughly validate the immense potential of tensor decomposition in handling high-dimensional mappings in continuous domains. 

\textbf{Tensor Decomposition for Operator Learning }
As operator learning is applied to increasingly complex scientific computing scenarios, controlling its parameter scale and memory footprint has become a recent research hotspot. 
To this end, researchers have begun exploring the integration of various tensor dimensionality reduction and compression techniques into advanced neural operator variants. 
In terms of frequency-domain transformation dimensionality reduction, the Factorized FNO (F-FNO) \citep{F-FNO} introduces fourier factorization to process each spatial dimension orthogonally and independently in the frequency domain. 
This successfully reduces model complexity by an order of magnitude while enabling stable scaling to deeper network structures. 
Following this dimensionality reduction paradigm, D-FNO \citep{D-FNO} decomposes high-dimensional latent representations into a series of rank-1 tensor products, converting 3D FFTs into 1D FFTs, thereby further reducing computational complexity. 
To address the parameter redundancy of global convolution kernels, architectures like Tucker-FNO \citep{Tucker-FNO} and MG-TFNO \citep{MG-TFNO} impose strict low-rank tensor constraints directly on the weight matrices, aiming to capture richer frequency-domain features while simultaneously reducing parameter counts. 
In the dimension of spatial basis extraction, POD-DeepONet \citep{POD-DeepONet} leverages proper orthogonal decomposition to pre-extract the dominant modes of the data as basis functions, significantly elevating the representational efficiency of the operator through effective spatial dimensionality reduction. 
Furthermore, the advantages of tensor decomposition have even extended to the optimization and solving stages of operator models. 
TensorGRaD \citep{TensorGRaD} decomposes gradients into the sum of low-rank and sparse tensors, efficiently capturing both global trends and high-frequency outliers in the evolutionary states of optimizers. 

\section{Proposed Methods} \label{sec3}
\subsection{The Setting of GNOs}
\label{Sec3}

Neural Operators learn solution operators associated with PDEs, mapping input functions to output functions in function spaces. This formulation supports mesh-independent, high-fidelity, and instantaneous inference after training. The detailed PDE configurations are deferred to Appendix \ref{PDEs}. 

Formally, let ${\cal A}$ and ${\cal U}$ be Banach spaces of functions defined on bounded domains $D \subset {\mathbb{R}^{{d_x}}}$, where ${\cal A}$ represents the input function space and ${\cal U}$ represents the objective function space. 
The purpose of operator learning is to approximate the ground-truth mapping $\mathcal{G}:\mathcal{A} \to \mathcal{U}$ by learning a neural operator ${\mathcal{G}_\theta }$ from a set of observations $\left\{ {{a_i},{u_i}} \right\}_{i = 1}^N$. 
The mapping ${\mathcal{G}_\theta }$ is defined as ${\mathcal{G}_\theta }:\mathcal{A} \to \mathcal{U},\theta  \in {\mathbb{R}^p}.$ We consider the loss function $L$ and learn neural operators by solving the problem
\begin{equation}
    \mathop {\min }\limits_\theta  \frac{1}{N}\sum\limits_{i = 1}^N {L\left( {{\mathcal{G}_\theta }\left( {{a_i}} \right),{u_i}} \right)}. 
\end{equation}
A standard Neural Operator (NO) framework consists of a lifting operator, iterative NO layers, and a projection operator. For $l = 1,2, \ldots ,L$, each NO layer updates the node feature $v^{(l)}$ via:  
\begin{equation}
    {v^{\left( {l + 1} \right)}}\left( x \right): = \sigma \left( {W^{\left( l \right)}}{v^{\left( l \right)}}\left( x \right) \right.   \left. + \bigl( {\mathcal{K}\left( {a;\phi } \right){v^{\left( l \right)}}} \bigr)\left( x \right) \right),\forall x=(x_1, ..., x_{d_x}) \in D, 
\end{equation}
where $\sigma $ is a nonlinear activation function, ${W^{(l)}}{v^{\left( l \right)}}\left( x \right)$ is the residual connection, and $\mathcal{K}$ represents the kernel integration operator, which is parameterized with $\phi$ by a neural network as 
\begin{equation}
    \left( {\mathcal{K}\left( {a;\phi } \right){v^{\left( l \right)}}} \right)\left( x \right): =  \int_D {\kappa \left( {x,y,a\left( x \right),a\left( y \right);\phi } \right){v^{\left( l \right)}}\left( y \right)dy} ,\forall x=(x_1, ..., x_{d_x}) \in D, 
\end{equation}
where $\kappa $ is an integral kernel. We assume that $dy$ defaults to the Lebesgue measure on ${\mathbb{R}^{{d_x}}}$. 

Consequently, NO can be expressed as 
\begin{equation}
    {\mathcal{G}_\theta }: = \mathcal{Q} \circ {\mathcal{L}^{\left( L \right)}} \circ {\mathcal{L}^{\left( {L - 1} \right)}} \circ  \cdots  \circ {\mathcal{L}^{\left( 1 \right)}} \circ \mathcal{P}. 
\end{equation}
Standard GNOs realize this continuous integration through message passing mechanism of the graph neural network \citep{GNN}. To handle unstructured grids and manage computational overhead, GNOs employ domain truncation (restricting integration to a local radius $r$) and the Nyström approximation \citep{Nystrom} (subsampling nodes for Monte-Carlo integration). These approximations fundamentally convert the continuous integral into discrete message-passing aggregations \citep{Gilmer2017} over a constructed graph topology. The extended mathematical formulations of these standard GNO approximation techniques are detailed in Appendix \ref{background_GNOs}. 

However, although GNOs effectively reduce the computational cost of global integration by introducing domain truncation and Nyström approximation, thereby enabling it to handle unstructured grid data, it still faces severe challenges when faced with high-dimensional input spaces or complex integration kernel modeling. 
The primary impediment pertains to the parameterization approach of its integral kernel $\kappa $. 
Traditional GNO relies on a monolithic architecture neural network such as MLP to approximate high-dimensional kernel functions. 
As the input dimension increases, the number of parameters in this network grows exponentially, which is known as the ``curse of dimensionality". 
Furthermore, during the spatial aggregation phase, the grid must explicitly construct and store the complete high-dimensional integral kernel tensor on every single edge. This edge-level dependency triggers an exponential explosion in parameter count and memory footprint, thereby exposing the fatal memory flaws of standard GNOs. 

\subsection{Functional Tensor Decomposition for GNO}
\label{Sec4}

As discussed above, the kernel integral operator in the standard GNO relies on a single, large neural network to parameterize the integral kernel. 
This design leads to substantial computational complexity. 
To address this issue, we incorporate functional tensor decomposition (FTD) into GNO and propose the FTD-GNO model. 
The core idea is to represent the high-dimensional, parameter-intensive kernel as a composition of low-dimensional, low-rank component functions. 
This method significantly reduces the number of parameters while preserving the dominant components of the tensor structure, thereby enabling effective extraction of key features from the physical field. 

\subsubsection{Functional Tensor Decomposition of the kernel}  \label{Sec4.1}

Let the generalized input to the kernel be denoted by the tuple $z = \left( {{z_1}, \ldots ,{z_d}} \right)$, where $d$ represents the number of physical input variable blocks. For our general integral kernel formulation, the input variables correspond to vector blocks of query point, the neighboring point, and their associated field values, i.e., $z = \left( {x,y,a\left( x \right),a\left( y \right)} \right)$, hence we have $d=4$. Assuming each variable block ${z_i}$ is located/evaluated across ${I_i}$ discrete elements, the fully instantiated integral kernel will essentially form a massive $d$-dimensional tensor of size ${I_1} \times {I_2} \times  \cdots  \times {I_d}$. Instead of using a single monolithic network to explicitly construct this dense tensor, we decouple the integral kernel $\kappa \left( {z;\Phi } \right)$ into a composition of $d$ smaller neural networks: 
\begin{equation}
    \kappa \left( {z;\Phi } \right) = {\text{Decomp}}\left( {{\Phi ^{\left( 1 \right)}}\left( {{z_1};{\theta _1}} \right), \ldots ,{\Phi ^{\left( d \right)}}\left( {{z_d};{\theta _d}} \right)} \right),
\end{equation}
where ${\Phi ^{\left( i \right)}}\left(  \cdot  \right)$ represents a univariate neural network processing the $i$-th variable block $z_i$ with learnable parameter ${\theta _i}$, and ${\text{Decomp}}\left(  \cdot  \right)$ denotes a classical tensor decomposition structure. 
Taking the full state $z = \left( {x,y,a\left( x \right),a\left( y \right)} \right)$ as the standard example: 
\begin{align}
    \kappa \left( {z;\Phi } \right) = {\text{Decomp}}\left( {{\Phi ^{\left( 1 \right)}}\left( {x;{\theta _1}} \right),{\Phi ^{\left( 2 \right)}}\left( {y;{\theta _2}} \right),{\Phi ^{\left( 3 \right)}}\left( {a\left( x \right);{\theta _3}} \right),{\Phi ^{\left( 4 \right)}}\left( {a\left( y \right);{\theta _4}} \right)} \right). \label{Eq9}
\end{align}
According to different PDEs, i.e. the dependence of $u$ on function  
$a$, it is necessary to select different types of integral kernel. 
The complete mathematical formulations for functional tensor decomposition of these general kernels are detailed in Appendix \ref{math of FTD-GNO}. 
In this work, we consider three established decomposition types in tensor theory, including CP \citep{CP}, TT \citep{TT} and Tucker decomposition \citep{Tucker}. 

\textbf{CP-GNO: }
The CP decomposition models the integral kernel as the sum of outer products of multiple rank-one functional tensors. The $d$ factor networks generate functional vectors, and the integral kernel is explicitly reconstructed as 
\begin{equation}
    \kappa \left( {{z_1}, \ldots ,{z_d};\Phi } \right) = \sum\limits_{r = 1}^R {A _r^{\left( 1 \right)}\left( {{z_1}} \right) \circ A _r^{\left( 2 \right)}\left( {{z_2}} \right) \circ  \cdots  \circ A _r^{\left( d \right)}} \left( {{z_d}} \right), 
\end{equation} 
where $A _r^{\left( i \right)}\left( {{z_i}} \right)$ explicitly denotes the continuous function of the $r$-th component for the $i$-th variable block.  
Here, $R$ is the rank of the CP decomposition and $ \circ $ denotes the outer product. 

\textbf{TT-GNO: }
The TT decomposition represents a tensor as a series of kernel tensors that are linked by chains. 
For $i = 1,2, \cdots ,d$, $d$ factor networks are used to generate core tensors, and then the integral kernel is reconstructed as 
\begin{equation}
   \kappa \left( {{z_1}, \ldots ,{z_d};\Phi } \right) = \sum\limits_{{r_1} = 1}^{{R_1}} {\sum\limits_{{r_2} = 1}^{{R_2}} { \cdots \sum\limits_{{r_{d - 1}} = 1}^{{R_{d - 1}}} {A _{{i_1},{r_0},{r_1}}^{\left( 1 \right)}\left( {{z_1}} \right)A _{{i_2},{r_1},{r_2}}^{\left( 2 \right)}\left( {{z_2}} \right) \cdots A _{{i_d},{r_{d - 1}},{r_d}}^{\left( d \right)}\left( {{z_d}} \right)} } }, 
\end{equation}
where ${R_i}$ represents the TT rank between modes $i$ and $i + 1$, and ${R_0} = {R_d} = 1$ is satisfied. 

\textbf{Tucker-GNO: }
The Tucker decomposition represents a tensor as the product of a core tensor and multiple factor matrices. 
Given Tucker rank $\{R_i\}^d_{i=1}$, the Tucker-GNO uses $d$ factor networks to generate the factor matrix for each mode. 
Then, the integral kernel is reconstructed as 
\begin{equation}
    \kappa \left( {{z_1}, \ldots ,{z_d};\Phi } \right) = \mathcal{C}{ \times _1}{A^{\left( 1 \right)}}\left( {{z_1}} \right){ \times _2}{A^{\left( 2 \right)}}\left( {{z_2}} \right) \cdots { \times _d}{A^{\left( d \right)}}\left( {{z_d}} \right), 
\end{equation}
where $\mathcal{C} \in {\mathbb{R}^{{R_1} \times {R_2} \times  \cdots  \times {R_d}}}$ is the learnable core tensor and ${ \times _i}$ represents the mode-$i$ tensor-matrix product. 

For the above GNO based on function tensor decomposition, these factor matrices/core tensors are learned and parameterized through $d$ small  
neural networks over their block variables, respectively. 
Subsequently, the reconstructed tensor, i.e. the integration kernel $\kappa \left( {z;\Phi } \right)$, is used in the integration operation of the kernel integration operator. 

\subsubsection{Model Architecture} \label{Sec4.2}

\begin{figure}
    \centering
    \includegraphics[width=0.91\linewidth]{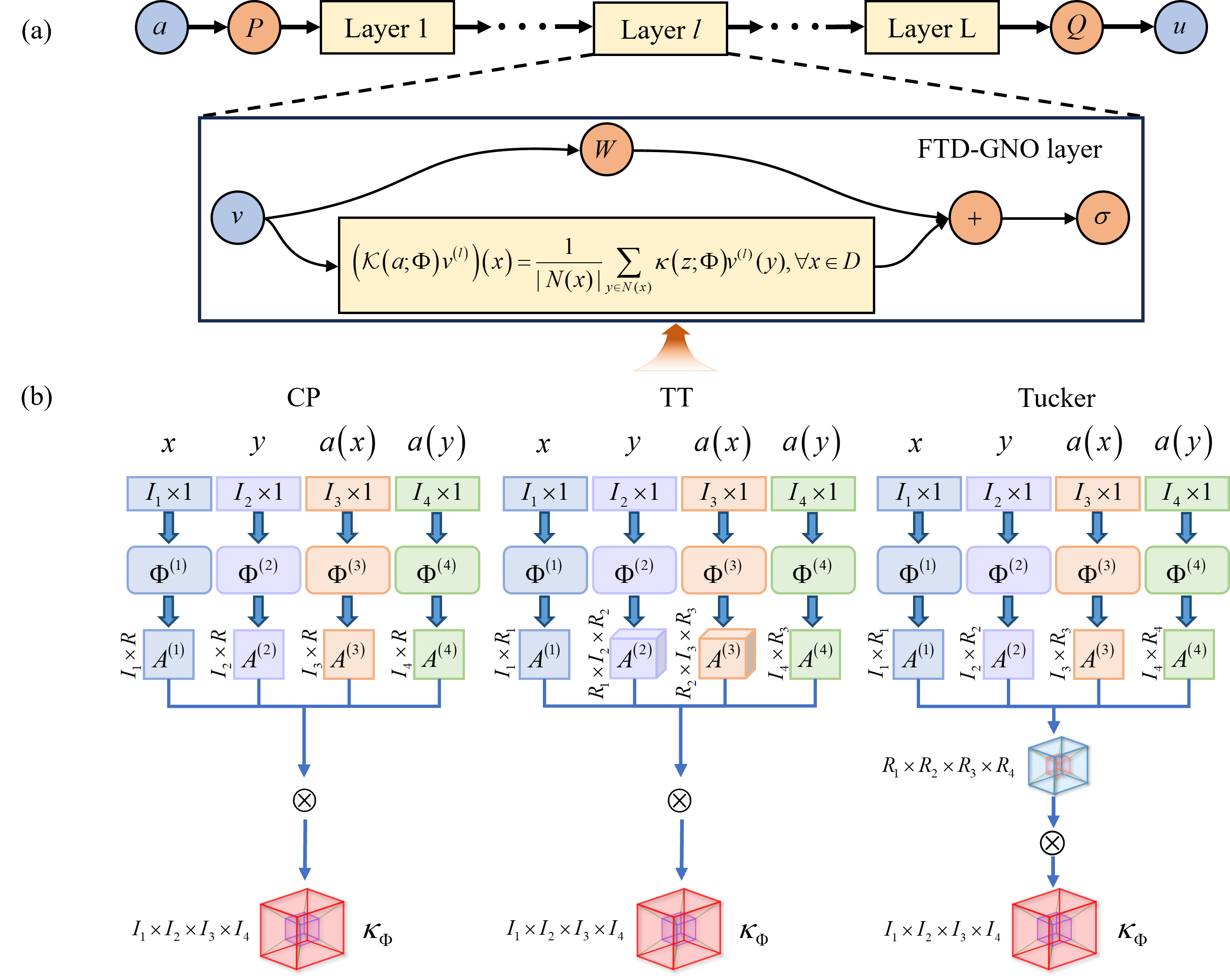}
    \caption{(a) The full architecture of proposed neural operator; (b) example of integral kernel parameterization with Functional Tensor Decompositions on a integral type of four variables. }
    \label{fig2}
\end{figure}

Our FTD-GNO adheres to the standard GNO architecture, which is 
\begin{equation}
    {\mathcal{G}_\theta }: = \mathcal{Q} \circ {\mathcal{L}^{\left( L \right)}} \circ {\mathcal{L}^{\left( {L - 1} \right)}} \circ  \cdots  \circ {\mathcal{L}^{\left( 1 \right)}} \circ \mathcal{P}, 
\end{equation}
where $\mathcal{P}$ is the lifting layer, $\mathcal{Q}$ is the projection layer, and each $\mathcal{L}^{(l)}$ is a neural operator layer. 
Here, we parameterize our integral kernel by $\kappa \left( {z;\Phi } \right)$. To explicitly distinguish from the standard GNOs formulation Section \ref{Sec3}, which uses $\phi$ to denote the parameter space of a single monolithic neural network, we employ $\Phi $ to represent the comprehensive collection of parameters from the $d$ decoupled factor networks and the learnable core tensor. And then we learn the mapping of neural operator layer via kernel integration operator parameterization 
\begin{equation}
    \left( {\mathcal{K}\left( {a;\Phi } \right){v^{\left( l \right)}}} \right)\left( x \right): = \int_D {\kappa \left( {z;\Phi } \right){v^{\left( l \right)}}\left( y \right)dy}\approx \frac{1}{{\left| {N\left( x \right)} \right|}}\sum\limits_{y \in N\left( x \right)} {\kappa \left( {z;\Phi } \right){v^{\left( l \right)}}\left( y \right)} ,\forall x \in D  
\end{equation}
and node feature propagation 
\begin{align}
    {v^{\left( {l + 1} \right)}}\left( x \right) : = \sigma \left( {W^{\left( l \right)}}{v^{\left( l \right)}}\left( x \right) \right.  \left. + \left( {\mathcal{K}\left( {a;\Phi } \right){v^{\left( l \right)}}} \right)\left( x \right) \right),\forall x \in D.  
\end{align}
It is worth noting that we borrow the same calculation method as GNO, completing the calculation of the above equation through domain truncation and Nyström approximation. 
As shown in Figure \ref{fig2}, we provide the full architecture of proposed neural operator as well as an example of integral kernel parameterization for (\ref{Eq9}) in three tensor decomposition modes. 

\subsubsection{Complexity Analysis} \label{Sec4.3}

This section aims to provide a comprehensive analysis of the computational efficiency by comparing standard GNO with the proposed FTD-GNO. 
The focus is on elucidating how FTD effectively mitigates the so-called ``curse of dimensionality" during the learning process of high-dimensional integral kernels. 
Assume that the single kernel network in the standard GNO has a width of $W$ and a depth of $L$. 
For FTD-GNO, its $d$ univariate subnetworks $\Phi^{(i)}$ have a width of $w$ and a depth of $l$.  

Under the standard network architecture, both the computational complexity and the parameter count for a single integral kernel evaluation in GNO are $O(W^2L)$. 
As the input dimension $d$ increases, the network width $W$ often needs to expand linearly or even exponentially to adequately capture high-dimensional feature interactions, thereby triggering a severe curse of dimensionality and significant computational redundancy. 
In contrast, FTD-GNO innovatively decouples the high-dimensional integral kernel into a composition of $d$ factor networks. 
The comprehensive theoretical complexity comparison is summarized in Table \ref{Tab1:complexity_analysis}. Detailed mathematical derivations for each tensor format (CP, TT, Tucker) and memory scaling proofs are provided in Appendix \ref{complexity analysis}. 

\begin{table*}[ht]
    \centering 
    \caption{Complexity analysis of different methods.}
    \label{Tab1:complexity_analysis}
    \begin{tabular}{l c c c}
        \toprule
        Method & Single Kernel Evaluation & Network Parameters & Memory Occupancy \\
        \midrule
        GNO        & $O(W^2L)$            & $O(W^2L)$          & $O(I^d)$ \\
        CP-GNO     & $O(dw^2lR)$          & $O(dw^2lR)$        & $O(Id)$ \\
        TT-GNO     & $O(d(w^2l + R^2))$   & $O(dw^2lR^2)$      & $O(Id)$ \\
        Tucker-GNO & $O(dw^2l + R^d)$     & $O(R^d + dw^2lR)$  & $O(Id)$ \\
        \bottomrule
    \end{tabular}
\end{table*}

As demonstrated in Table \ref{Tab1:complexity_analysis}, FTD-GNO achieves an effective separation of variables. 
The divide-and-conquer approach is demonstrably advantageous in terms of model efficiency, as listed below: 

\textbf{1. Optimizing Memory Occupancy:} Leveraging independent feature decomposition and integral operator reconstruction, the model reduces the memory footprint of input construction to a linear scale, unlocking scalable training for high-resolution physical fields. 

\textbf{2. Reducing Computational Complexity:} The model has been demonstrated to reduce the computational cost of a single kernel evaluation from a heavy polynomial scale to a linear scale, thereby substantially enhancing the forward inference efficiency of the operator. 

\textbf{3. Slashing Parameter Scale:} By avoiding redundant network expansion, the model preserves equivalent expressive capacity while inherently mitigating overfitting risks, thereby facilitating the lightweight deployment of large-scale operators. 

\section{Experiments}
\label{Sec5}

We evaluate our proposed models on the Darcy flow, Navier-Stokes (NS) sourced from the NeuralOperator library and elasticity on sample-dependent irregular meshes.
Appendix \ref{experimental_settings} reports the detailed experimental settings and baseline comparisons. 

\subsection{Main Comparisons} \label{Baseline Comparison}

\begin{table*}[t]
    \caption{Comprehensive benchmark of relative $L_2$ errors. Peak memory is reported in MB. }
    \label{Tab:more baseline}
    \centering 
    \begin{tabular}{l c c c c c c c}
        \toprule
        \multirow{2}{*}{\makecell{Model}} & \multicolumn{2}{c}{Darcy $16\times16$} & \multicolumn{2}{c}{NS $16\times16$} & \multicolumn{2}{c}{NS $128\times128$} \\
        \cmidrule(lr){2-3} \cmidrule(lr){4-5} \cmidrule(lr){6-7}
          &  Error & Peak memory &  Error & Peak memory & Error & Peak memory \\
        \midrule
        GNO        & 0.1928 & 435.43 & 0.5482 & 420.61 & OOM & OOM \\
        CP-GNO     & 0.2114 & 197.02 & 0.5538 & 181.38 & 0.5510 & 6002.06 \\
        TT-GNO     & 0.2114 & 197.12 & 0.5538 & 181.38 & 0.5510 & 6009.94  \\
        Tucker-GNO & 0.2104 & 201.68 & 0.5493 & 183.32 & 0.5489 & 8963.39 \\
        \midrule
        FNO        & 0.1042 & 93.58 & 0.0789 & 92.92 & 0.0783 & 1934.74 \\
        FNOGNO     & 0.1024 & 195.93 & 0.1152 & 276.77 & OOM & OOM \\
        CP-FNOGNO  & 0.1026 & 122.80 & 0.1156 & 123.37 & 0.1203 & 6104.55 \\
        TT-FNOGNO  & 0.1026 & 122.55 & 0.1156 & 123.00 & 0.1203 & 6207.68 \\
        Tucker-FNOGNO  & 0.1067 & 124.80 & 0.1163 & 128.49 & 0.1146 & 8469.61 \\
        \bottomrule
    \end{tabular}
\end{table*}
\begin{table}[t]
    \caption{Experiment results of relative $L_2$ errors, training time, and parameters on Darcy flow. }
    \label{Tab2:main results}
    \centering 
    \begin{tabular}{l c c c c c c c c c}
        \toprule
        \multirow{2}{*}{\makecell{Model}} & \multicolumn{3}{c}{$16\times16$} & \multicolumn{3}{c}{$32\times32$} & \multirow{2}{*}{\makecell{Parameters}} & Parameters \\
        \cmidrule(lr){2-4} \cmidrule(lr){5-7}
          &  Error & Time (s) &  Speedup & Error & Time (s) &  Speedup & & Reduction \\
        \midrule
        GNO        & 0.1872 & 6542.28 & -- & 0.1723 & 22602.04 & -- & 95265 & --  \\
        CP-GNO     & 0.1975 & 647.00 & \textbf{10.11$\times$} & 0.1815 & 545.27 & \textbf{41.45$\times$} &  74725 &  \textbf{21.56\%} \\
        TT-GNO     & 0.1975 & 534.03 & \textbf{12.25$\times$} & 0.1815 & 654.29 & \textbf{34.54$\times$} & 74725  & \textbf{21.56\%}  \\
        Tucker-GNO & 0.2157 & 665.17 & \textbf{9.84$\times$} & 0.2214 & 635.18 & \textbf{35.58$\times$} & 72709 & \textbf{23.68\%} \\
        \bottomrule
    \end{tabular}
\end{table}

\textbf{Extensibility: Whether FTD benefits different operator architectures.} 

We consider two groups of controlled comparisons: GNO against CP-, TT-, and Tucker-GNO, and FNOGNO against CP-, TT-, and Tucker-FNOGNO. FNOGNO replaces the preceding graph operator layers with Fourier layers while retaining the final graph integral. Its factorized variants modify only that final kernel. As shown in Table~\ref{Tab:more baseline}, FTD reduces peak memory in both groups, extending its benefit from graph-only to Fourier-graph architectures. At higher NS resolution, the factorized variants remain trainable when FNOGNO encounter OOM. Because FTD modifies the kernel parameterization and contraction locally, it can be incorporated into other operator frameworks containing graph-integral layers that admit the proposed factorization. 

\textbf{Effectiveness: Whether FTD reduces graph-kernel costs beyond weight compression.}

As shown in Table~\ref{Tab:LR-GNO} (Appendix ~\ref{sec:E.2.1}), LR-GNO and FTD-GNO exhibit substantially different memory costs despite closely matched parameter counts. LR-GNO compresses the kernel MLP weights but preserves its evaluation on neighboring node pairs, whereas FTD reorganizes kernel evaluation and aggregation through reusable node-wise factors. This comparison indicates that parameter reduction alone does not explain the memory advantage and supports the role of the factorized contraction. The slightly lower error of LR-GNO also shows that matching parameter counts does not imply identical approximation or optimization behavior. 

\textbf{Benchmarks: Whether FTD remains effective in different PDE tasks.} 

For regular grid benchmarks, FTD reduces peak memory relative to the corresponding unfactorized baselines in both architecture groups, shown in Table \ref{Tab:more baseline}. Factorization also enables higher-resolution NS training when the unfactorized baselines encounter OOM. Table \ref{Tab2:main results} reports that the proposed models achieve substantial training speedups and significant parameter reductions. However, graph-only configurations exhibit relatively high errors in NS, where FNO provides more accurate predictions. Incorporating Fourier operator layers improves the predictive performance of operator models. Accordingly, we view FTD as a memory-efficient component for graph integral architectures, rather than establishing FTD-GNO as a uniformly superior operator across PDE tasks. 

For the irregular benchmark, under the same training setting, all three FTD-GNO variants reduce peak memory and training time relative to GNO, shown in Table~\ref{Tab:irregular} (Appendix \ref{sec:E.2.1}). This demonstrates that the efficiency benefit extends to irregular meshes, where the graph connectivity varies across samples. Moreover, the additional 300-epoch runs improve the factorized models beyond the error of the 100-epoch GNO while retaining lower memory and recorded total runtime. These results show that the reduced training cost can be used to support additional optimization. 

In summary, these results consistently support FTD as an effective approach for reducing graph-integral costs. 
Further analyses are provided in Appendix \ref{sec:E.2.1}. Moreover, Appendix \ref{sec:E.2.2} reports the effects of rank and model capacity (Tables~\ref{Tab:rank1} and~\ref{Tab:rank2}), 
as well as the effects of kernel input variable blocks on peak memory (Table \ref{Tab5:peak_memory2}). 

\subsection{Memory Footprint}

Tables \ref{tab:peak_memory_main} and \ref{tab:peak_memory_summary} compare peak memory across resolutions and neighborhood radii. GNO memory increases sharply as the graph becomes denser and reaches OOM at the $64 \times 64$ resolutions. 
Conversely, all FTD-GNO variants adhere strictly to a stable, linear growth trajectory. By algebraically decoupling computation from the edge dimension, FTD-GNO demonstrates remarkable memory invariance to radius expansion and ensures memory proportionality solely to the total node count, thereby completely shattering the hardware barriers for high-resolution physical field modeling.

\begin{table*}[t]
\centering
\caption{Peak GPU memory footprint (MB) of different models under varying resolutions and radii.}
\label{tab:peak_memory_main}
\setlength{\tabcolsep}{3.3pt} 
\begin{tabular}{lccccccccc}
\toprule
& \multicolumn{3}{c}{$16 \times 16$}
& \multicolumn{3}{c}{$32 \times 32$}
& \multicolumn{3}{c}{$64 \times 64$} \\
\cmidrule(lr){2-4} \cmidrule(lr){5-7} \cmidrule(lr){8-10}
Model
& $r{=}0.08$ & $r{=}0.12$ & $r{=}0.16$
& $r{=}0.08$ & $r{=}0.12$ & $r{=}0.16$
& $r{=}0.08$ & $r{=}0.12$ & $r{=}0.16$ \\
\midrule
GNO
& 420.6 & 696.3 & 1511.0
& 6426.6 & 13208.9 & 19733.3
& \textsc{OOM} & \textsc{OOM} & \textsc{OOM} \\

CP-GNO ($R{=}8$)
& 182.7 & 182.7 & 182.9
& 680.3 & 682.0 & 683.6
& 2688.8 & 2715.9 & 2753.5 \\
CP-GNO ($R{=}16$)
& 274.7 & 274.8 & 275.0
& 1048.4 & 1050.0 & 1051.6
& 4160.8 & 4187.4 & 4225.5 \\
CP-GNO ($R{=}32$)
& 458.8 & 458.9 & 459.1
& 1784.5 & 1786.1 & 1787.7
& 7104.9 & 7132.0 & 7168.9 \\

TT-GNO ($R{=}8$)
& 182.7 & 182.7 & 182.9
& 680.3 & 682.0 & 683.6
& 2688.9 & 2715.9 & 2753.5 \\
TT-GNO ($R{=}16$)
& 274.7 & 274.8 & 275.0
& 1048.4 & 1050.0 & 1051.6
& 4160.8 & 4187.4 & 4225.5 \\
TT-GNO ($R{=}32$)
& 458.8 & 458.9 & 459.1
& 1784.5 & 1786.1 & 1787.7
& 7104.9 & 7132.0 & 7168.9 \\

Tucker-GNO ($R{=}8$)
& 183.7 & 183.7 & 183.9
& 684.4 & 686.0 & 687.6
& 2706.8 & 2731.9 & 2769.9 \\
Tucker-GNO ($R{=}16$)
& 276.7 & 276.8 & 277.0
& 1056.4 & 1058.0 & 1059.6
& 4192.9 & 4218.8 & 4257.5 \\
Tucker-GNO ($R{=}32$)
& 462.8 & 462.9 & 463.1
& 1800.5 & 1802.2 & 1803.7
& 7168.9 & 7196.1 & 7233.0 \\
\bottomrule
\end{tabular}
\end{table*}
\begin{table*}[t]
\centering
\caption{Derived memory scaling and efficiency analysis from Table~\ref{tab:peak_memory_main}.}
\label{tab:peak_memory_summary} 
\small
\begin{tabular}{p{2.8cm}cccccccccc}
\toprule
Metric
& GNO
& CP-8 & CP-16 & CP-32
& TT-8 & TT-16 & TT-32
& Tcr-8 & Tcr-16 & Tcr-32 \\
\midrule 
$16{\times}16\,{\rightarrow}\,32{\times}32$, $r{=}0.12$
& 19.0$\times$
& 3.7$\times$ & 3.8$\times$ & 3.9$\times$
& 3.7$\times$ & 3.8$\times$ & 3.9$\times$
& 3.7$\times$ & 3.8$\times$ & 3.9$\times$ \\
$r{:}\,\,0.08\,{\rightarrow}\,0.16$, $32{\times}32$
& 3.1$\times$
& 1.0$\times$ & 1.0$\times$ & 1.0$\times$
& 1.0$\times$ & 1.0$\times$ & 1.0$\times$
& 1.0$\times$ & 1.0$\times$ & 1.0$\times$ \\

Reduction vs.\ GNO
($32\!\times\!32$, $r{=}0.16$)
& 1.0$\times$
& \textbf{28.9$\times$} & \textbf{18.8$\times$} & \textbf{11.0$\times$}
& \textbf{28.9$\times$} & \textbf{18.8$\times$} & \textbf{11.0$\times$}
& \textbf{28.7$\times$} & \textbf{18.6$\times$} & \textbf{10.9$\times$} \\
\bottomrule
\end{tabular}
\end{table*}

\subsection{Zero-Shot Super-Resolution}

\begin{wraptable}{r}{0.36\columnwidth}
    \vspace{-0.6em}
    \centering
    \setlength{\tabcolsep}{4pt} 
    \caption{$l_2$ relative test errors.}  
    \label{Tab3:Zero-Shot}
    \begin{tabular}{lcc}
        \toprule
        Model & $16\times16$ & $32\times32$ \\
        \midrule
        GNO     & 0.1872 & 0.1838 \\
        CP-GNO     & 0.1975 & 0.1956 \\
        TT-GNO     & 0.1975 & 0.1956 \\
        Tucker-GNO & 0.2157 & 0.2214 \\
        \bottomrule
    \end{tabular}
    \vspace{-0.8em}
\end{wraptable}

Resolution invariance enables neural operators trained on low-resolution grids to perform zero-shot inference on unseen higher-resolution grids without retraining. 
To assess zero-shot super-resolution, we train FTD-GNO at $16\times16$ and directly evaluate it at $16\times16$ and $32\times32$, shown in Table \ref{Tab3:Zero-Shot}. 
The results reveal that the relative test errors for all models remained within the same order of magnitude, which suggests the robust zero-shot super-resolution generalization capability of the proposed architecture. 

\section{Conclusion}
\label{Sec6}

In this paper, we propose FTD-GNO, a novel Graph Neural Operator architecture that reconstructs the integral kernel via functional tensor decomposition. 
Through the implementation of variable decoupling and algebraic integral reconstruction methodologies, we have successfully overcome the computational and memory limitations inherent to conventional GNOs when processing high-resolution physical fields. 
As a result, the memory requirement scales linearly and smoothly with the node size. 
Moreover, a systematic evaluation of three core variants is conducted, utilising CP, TT, and Tucker decompositions as the underlying methodologies. 
Across the regular-grid and irregular-mesh tasks, FTD reduces peak memory relative to corresponding unfactorized graph-integral baselines and also benefits a Fourier–graph hybrid. 
This breakthrough paves a highly efficient path for the engineering deployment of GNOs in large-scale, high-fidelity scientific computing. 
Appendix \ref{limitations} further discusses the limitations of FTD-GNO and outlines directions for future research. 

\bibliographystyle{unsrt}  

\clearpage
\appendix
\section{General Framework of Parametric PDEs} \label{PDEs}
This section delineates the general mathematical framework of parametric PDEs to establish the physical context for the operator learning discussed in Section \ref{Sec3}. 

In scientific computing and engineering, the evolution of complex physical systems is typically governed by PDEs. 
When variable parameters are introduced to account for uncertainty quantification or physical inverse problems, these parameter-driven equations are collectively termed parametric PDEs. 
Mathematically, let $D \subset \mathbb{R}^{d_x}$ denote a bounded spatial domain with boundary $\partial D$. 
We consider a general class of parametric PDEs formulated as: 
\begin{equation}
    \begin{aligned}
        (\mathcal{L}_a u)(x) &= f(x), && x \in D \\
        u(x) &= 0, && x \in \partial D
    \end{aligned}
\end{equation}
where $a(x)$ denotes the input parameter field of the system (e.g., physical properties of heterogeneous media, conductivity coefficients, or fluid viscosity), and $u(x)$ is the unknown target physical response field. 
The operator $\mathcal{L}_a$, which explicitly depends on the parameter field $a$, mathematically characterizes the underlying conservation laws and spatial differential rules governing the physical system. 
Meanwhile, $f(x)$ represents the source term or forcing function, corresponding to external driving forces (e.g., external energy sources, mass sinks, or body forces) acting within the interior of domain $D$. 
The aforementioned formulation is also subject to homogeneous Dirichlet boundary conditions. 

To be more specific, taking the steady-state Darcy flow, a prototypical fluid-in-porous-media problem widely benchmarked in the operator learning, can be governed by the following second-order elliptic PDE: 
\begin{equation}
    \begin{aligned}
        -\nabla \cdot (a(x) \nabla u(x)) &= f(x), && x \in D \\
        u(x) &= 0, && x \in \partial D.
    \end{aligned}
\end{equation}
Within this physical context, the input parameter $a(x)$ generally represents a highly heterogeneous permeability field within the medium, with the target solution $u(x)$ corresponding to the resulting fluid pressure field. 
Assuming the well-posedness of the PDE, for a fixed source term $f(x)$ and given boundary conditions, each specific parameter instance $a$ in the input parameter space is associated with to a unique physical response solution $u$. 
This strict dependence, inherently dictated by the underlying physical conservation laws, essentially establishes a nonlinear mapping defined between infinite-dimensional function spaces, i.e., $a \mapsto u$. 

\section{Extended Background on Standard GNOs} \label{background_GNOs}
This section details the domain truncation and Nyström approximation techniques utilized by standard Graph Neural Operators (GNOs) to discretize the continuous integral kernel operator described in Section \ref{Sec3}. 

\textbf{Domain truncation.} 
The spatial domain $D$ is discretised by $J$ points. 
Each discrete point, designated $x \in D$, is regarded as a node of the directed graph, and the node feature ${v^{\left( l \right)}}\left( x \right)$ is assigned to it. 
The edge connectivity is determined by the Lebesgue measure. 
Let $s:D \to \mathcal{B}\left( D \right)$ be the mapping from the point of $D$ to the Lebesgue measurable subsets of $D$, denoted as $\mathcal{B}\left( D \right)$. 
In general, we consider $s\left( x \right) = B\left( {x,r} \right) \cap D$, where $B\left( {x,r} \right) = \left\{ {y \in {\mathbb{R}^{d_x}}:{{\left\| {y - x} \right\|}_{{\mathbb{R}^{d_x}}}} < r,r > 0} \right\}$. 
Each node $x$ is connected to the node set located in $B\left( {x,r} \right)$ and is defined as the neighbourhood set $N\left( x \right)$. 
And for each neighbor $y \in N\left( x \right)$, we assign the edge weight $\left( {x,y,a\left( x \right),a\left( y \right)} \right)$. 
Therefore, the integral equation becomes 
\begin{align}
    \left( {\mathcal{K}\left( {a;\phi } \right){v^{\left( l \right)}}} \right)\left( x \right) =   \int_{B\left( {x,r} \right)} {\kappa \left( {x,y,a\left( x \right),a\left( y \right);\phi } \right){v^{\left( l \right)}}\left( y \right)dy} ,\forall x \in D. \label{Eq6}
\end{align}
This truncation is interpreted as constructing a local structure in the integral kernel $\kappa $. 
This local structure facilitates enhanced computational efficiency while remaining invariant to mesh-refinement. 
Indeed, given that the neighbourhood radius $r$ is selected in physical space, this is a pivotal characteristic that renders GNO mesh-independent. 
In addition, if the cost of computing original integral equation is $O( {{J^2}} )$, then utilising the selection of $s( x ) = B( {x,r} ) \cap D$ and assuming $D = {[ {0,1} ]^{d_x}}$, the complexity of computing (\ref{Eq6}) is approximately $O( {{r^{d_x}}{J^2}} )$. 
In fact, for the general case where the size of the set $s( x )$ is less than $D$, we always have the cost of calculating (\ref{Eq6}) as $O( {{c_s}{J^2}} )$, where ${c_s}$ is a constant that depends on $s$ and ${c_s} < 1$. 
This may significantly impact the actual calculation, so domain truncation is an effective means of reducing the cost of calculation. 

\textbf{Nyström approximation theorem.} 
Nyström approximation is a simple and effective method commonly used to calculate the kernel integral operator, which is mainly used to solve the problem of large computational overhead of integrating over the entire domain $D$. 
Its main idea is to randomly sample the points over which we compute the output function $u$. 
Specifically, if we uniformly sample $J' \ll J$ nodes from the original graph, the computational cost decreases from $O( {{J^2}} )$ to $O( {{{J'}^2}} )$, which is only related to the number of subsampled points $J'$. 
It is evident that the number of subsampled points selected is typically much smaller than the original discretisation points. 
Consequently, this results in a substantial reduction in the final computational cost. 

In accordance with the aforementioned definition, the message passing algorithm proposed by \citet{Gilmer2017} is utilised to construct a graph for the spatial domain $D$ of partial differential equations. 
The kernel integral is regarded as the aggregation of messages. 
Consequently, the message passing kernel integral operator with average aggregation is designated as 
\begin{align}
    \left( {{\cal K}\left( {a;\phi } \right){v^{\left( l \right)}}} \right)\left( x \right) =  \frac{1}{{\left| {N\left( x \right)} \right|}}\sum\limits_{y \in N\left( x \right)} {\kappa \left( {x,y,a\left( x \right),a\left( y \right);\phi } \right){v^{\left( l \right)}}\left( y \right)} ,\forall x \in D. \label{Eq7}
\end{align}
Here, (\ref{Eq7}) can be regarded as the Monte-Carlo approximation of the integral (\ref{Eq6}). 

\section{Mathematical Foundations of FTD-GNO: Decomposition} \label{math of FTD-GNO}

Table \ref{decomposition} details the explicit mathematical formulations of the Functional Tensor Decomposition applied to the integral kernel within FTD-GNO. Unlike the standard GNO, we explicitly expand the continuous integral kernel into a product of multiple low-rank tensor factors. This factorization is a fundamental prerequisite for achieving high computational efficiency. The table systematically illustrates the decomposition process as the input variables expand from basic spatial coordinates to a full physical state. It demonstrates how three mainstream tensor decomposition mechanisms (CP, TT, and Tucker) decouple the multi-dimensional, coupled kernel function into independent factor functions, which are parameterized by neural networks. Through this explicit functional expansion, the complex interactions among multiple variables are mapped into an implicit feature space of rank $R$. Specifically for the Tucker decomposition, a learnable core tensor is introduced to capture the non-linear coupling relationships across different rank dimensions. 
\begin{table}[t]
\centering
\caption{Functional tensor decomposition formulas of the integral kernel under CP, TT, and Tucker formats for multi-variable inputs. }
\label{decomposition} 
\small
\resizebox{\textwidth}{!}{
\begin{tabular}{lll}
\toprule
Type
& Standard GNO Formulation
& FTD-GNO Decomposition \\
\midrule
\multirow{6}{*}{\makecell[c]{CP}}
& ${\kappa \left( {x,y;\phi } \right)} $
& ${\sum\limits_{r = 1}^R {{A _r^{(1)}}\left( x \right){A _r^{(2)}}\left( y \right)} }$ \\

& ${\kappa \left( {x,y;\phi } \right){v^{\left( l \right)}}\left( y \right)} $
& ${\sum\limits_{r = 1}^R {{A _r^{(1)}}\left( x \right){A _r^{(2)}}\left( y \right){v^{\left( l \right)}}\left( y \right)} }$ \\

& ${\kappa \left( {x,y,a\left( y \right);\phi } \right)} $
& ${\sum\limits_{r = 1}^R {{A _r^{(1)}}\left( x \right){A _r^{(2)}}\left( y \right){A _r^{(3)}}\left( {a\left( y \right)} \right)} }$ \\

& ${\kappa \left( {x,y,a\left( y \right);\phi } \right){v^{\left( l \right)}}\left( y \right)} $
& ${\sum\limits_{r = 1}^R {{A _r^{(1)}}\left( x \right){A _r^{(2)}}\left( y \right){A _r^{(3)}}\left( {a\left( y \right)} \right){v^{\left( l \right)}}\left( y \right)} }$ \\

& ${\kappa \left( {x,y,a\left( x \right),a\left( y \right);\phi } \right)} $
& ${\sum\limits_{r = 1}^R {{A _r^{(1)}}\left( x \right){A _r^{(2)}}\left( y \right){A _r^{(3)}}\left( {a\left( x \right)} \right){A _r^{(4)}}\left( {a\left( y \right)} \right)} }$ \\

& ${\kappa \left( {x,y,a\left( x \right),a\left( y \right);\phi } \right){v^{\left( l \right)}}\left( y \right)} $
& ${\sum\limits_{r = 1}^R {{A _r^{(1)}}\left( x \right){A _r^{(2)}}\left( y \right){A _r^{(3)}}\left( {a\left( x \right)} \right){A _r^{(4)}}\left( {a\left( y \right)} \right){v^{\left( l \right)}}\left( y \right)} }$ \\
\midrule

\multirow{6}{*}{\makecell[c]{TT}}
& ${\kappa \left( {x,y;\phi } \right)} $
& ${\sum\limits_{r = 1}^R {{A _r^{(1)}}\left( x \right){A _r^{(2)}}\left( y \right)} }$ \\

& ${\kappa \left( {x,y;\phi } \right){v^{\left( l \right)}}\left( y \right)} $
& ${\sum\limits_{r = 1}^R {{A _r^{(1)}}\left( x \right){A _r^{(2)}}\left( y \right){v^{\left( l \right)}}\left( y \right)} }$ \\

& ${\kappa \left( {x,y,a\left( y \right);\phi } \right)} $
& ${\sum\limits_{{r_1},{r_2}} {{A _{{r_1}}^{(1)}}\left( x \right){A _{{r_1},{r_2}}^{(2)}}\left( y \right){A _{{r_2}}^{(3)}}\left( {a\left( y \right)} \right)} } $ \\

& ${\kappa \left( {x,y,a\left( y \right);\phi } \right){v^{\left( l \right)}}\left( y \right)} $
& ${\sum\limits_{{r_1},{r_2}} {{A _{{r_1}}^{(1)}}\left( x \right){A _{{r_1},{r_2}}^{(2)}}\left( y \right){A _{{r_2}}^{(3)}}\left( {a\left( y \right)} \right){v^{\left( l \right)}}\left( y \right)} }$ \\

& ${\kappa \left( {x,y,a\left( x \right),a\left( y \right);\phi } \right)} $
& ${\sum\limits_{{r_1},{r_2},{r_3}} {{A _{{r_1}}^{(1)}}\left( x \right){A _{{r_1},{r_2}}^{(2)}}\left( y \right){A _{{r_2},{r_3}}^{(3)}}\left( {a\left( x \right)} \right){A _{{r_3}}^{(4)}}\left( {a\left( y \right)} \right)} }$ \\

& ${\kappa \left( {x,y,a\left( x \right),a\left( y \right);\phi } \right){v^{\left( l \right)}}\left( y \right)} $
& ${\sum\limits_{{r_1},{r_2},{r_3}} {{A _{{r_1}}^{(1)}}\left( x \right){A _{{r_1},{r_2}}^{(2)}}\left( y \right){A _{{r_2},{r_3}}^{(3)}}\left( {a\left( x \right)} \right){A _{{r_3}}^{(4)}}\left( {a\left( y \right)} \right){v^{\left( l \right)}}\left( y \right)} }$ \\
\midrule

\multirow{6}{*}{\makecell[c]{Tucker}}
& ${\kappa \left( {x,y;\phi } \right)} $
& ${\sum\limits_{{r_1},{r_2}} {{\mathcal{C}_{{r_1},{r_2}}}{A _{{r_1}}^{(1)}}\left( x \right){A _{{r_2}}^{(2)}}\left( y \right)} }$ \\

& ${\kappa \left( {x,y;\phi } \right){v^{\left( l \right)}}\left( y \right)} $
& ${\sum\limits_{{r_1},{r_2}} {{{\cal C}_{{r_1},{r_2}}}{A _{{r_1}}^{(1)}}\left( x \right){A _{{r_2}}^{(2)}}\left( y \right){v^{\left( l \right)}}\left( y \right)} }$ \\

& ${\kappa \left( {x,y,a\left( y \right);\phi } \right)} $
& ${\sum\limits_{{r_1},{r_2},{r_3}} {{\mathcal{C}_{{r_1},{r_2},{r_3}}}{A _{{r_1}}^{(1)}}\left( x \right){A _{{r_2}}^{(2)}}\left( y \right){A _{{r_3}}^{(3)}}\left( {a\left( y \right)} \right)} }$ \\

& ${\kappa \left( {x,y,a\left( y \right);\phi } \right){v^{\left( l \right)}}\left( y \right)} $
& ${\sum\limits_{{r_1},{r_2},{r_3}} {{{\cal C}_{{r_1},{r_2},{r_3}}}{A _{{r_1}}^{(1)}}\left( x \right){A _{{r_2}}^{(2)}}\left( y \right){A _{{r_3}}^{(3)}}\left( {a\left( y \right)} \right){v^{\left( l \right)}}\left( y \right)} }$ \\

& ${\kappa \left( {x,y,a\left( x \right),a\left( y \right);\phi } \right)} $
& ${\sum\limits_{{r_1},{r_2},{r_3},{r_4}} {{\mathcal{C}_{{r_1},{r_2},{r_3},{r_4}}}{A _{{r_1}}^{(1)}}\left( x \right){A _{{r_2}}^{(2)}}\left( y \right){A _{{r_3}}^{(3)}}\left( {a\left( x \right)} \right){A _{{r_4}}^{(4)}}\left( {a\left( y \right)} \right)} }$ \\

& ${\kappa \left( {x,y,a\left( x \right),a\left( y \right);\phi } \right){v^{\left( l \right)}}\left( y \right)} $
& ${\sum\limits_{{r_1},{r_2},{r_3},{r_4}} {{{\cal C}_{{r_1},{r_2},{r_3},{r_4}}}{A _{{r_1}}^{(1)}}\left( x \right){A _{{r_2}}^{(2)}}\left( y \right){A _{{r_3}}^{(3)}}\left( {a\left( x \right)} \right){A _{{r_4}}^{(4)}}\left( {a\left( y \right)} \right){v^{\left( l \right)}}\left( y \right)} }$ \\
\bottomrule
\end{tabular}}
\end{table}

\textbf{Proposition C.1 (increasing-capacity kernel approximation).} Let $\mathcal{Z}_i$ be compact subsets of Euclidean spaces and let $\kappa \in C(\mathcal{Z}_1 \times \cdots \times \mathcal{Z}_d; \mathbb{R})$. For every $\varepsilon > 0$, there are a finite rank $R$ and continuous functions $A_r^{(i)} \in C(\mathcal{Z}_i; \mathbb{R})$ such that
\begin{align}
\left\| \kappa(z_1, \ldots, z_d) - \sum_{r=1}^R \prod_{i=1}^d A_r^{(i)}(z_i) \right\|_\infty < \varepsilon.
\end{align}
If the factor-network family is uniformly dense in continuous functions on each $\mathcal{Z}_i$, the factors can be replaced by sufficiently capable networks with the same conclusion after increasing their capacities as necessary. The result extends component-wise to a finite-dimensional channel-wise kernel. 

\textbf{Proof.}
Consider the set $\mathcal{A}$ of finite sums of products $\sum_{r=1}^R \prod_i f_{ir}(z_i)$, with continuous block functions $f_{ir}$. This set contains constants, is closed under addition and multiplication, and therefore forms a unital subalgebra of the continuous functions on the compact product domain. It separates points: if two tuples differ in a block, a continuous function on that block separates them, while constants on the remaining blocks complete a product function. The real Stone--Weierstrass theorem implies that $\mathcal{A}$ is uniformly dense. Choose a finite expansion with $d$ many bounded factors, the product perturbation inequality
\begin{align}
\left\| \prod_{i=1}^d f_i - \prod_{i=1}^d g_i \right\| \le \sum_{j=1}^d |f_j - g_j| \prod_{i<j} |g_i| \prod_{i>j} |f_i|
\end{align}
allows each factor to be approximated by a network with a sufficiently small uniform error so that the total finite-sum perturbation is below $\varepsilon/2$. The triangle inequality proves the result. TT can realize a CP sum using diagonal intermediate cores, and Tucker using a superdiagonal core, so both inherit this increasing-rank approximation property. For finitely many channels, apply the argument to each component and pad ranks as needed. 
This proposition is consistent with functional tensor approximation results used by \citet{F-INR, PINN}. 

\section{Detailed Complexity and Memory Analysis} \label{complexity analysis}
The appendix provides the rigorous mathematical derivations for the complexity analysis summarized in Section \ref{Sec4.3} (Table \ref{Tab1:complexity_analysis}). 

\subsection{Computational and Parameter Complexity}
Specifically, the standard GNO integral kernel requires $\mathcal{O}(W^2L)$ for both computation and parameter count. FTD-GNO decomposes this into $d$ factor networks. The specific complexities for the three variants are derived as follows: 

\textbf{CP-GNO} models the integral kernel as a composition of factors with rank $R$. Its $d$ factor networks have a forward propagation complexity of $O(dw^2l)$, and the outer product composition requires a computational complexity of $O(dR)$. Thus, the overall computational cost for a single kernel evaluation is $O\left( {d{w^2}lR} \right)$, with the total network parameter count identically being $O(dw^2lR)$. 

\textbf{TT-GNO} leverages the chain-like contraction property of the tensor train structure to elegantly circumvent the explicit construction of full high-dimensional tensors. The boundary TT-ranks are set to $R_0=R_d=1$, and for computational simplicity, the intermediate ranks are assumed to be $R_i=R$. The forward propagation complexity of this architecture is $O(dw^2l)$, and the tensor contraction complexity is $O(dR^2)$. Consequently, its total computational cost for a single kernel evaluation is $O(d(w^2l+R^2))$. With regard to the parameter count, the first and last networks have an output dimension of $R$, contributing $O({w^2}lR)$ parameters; the intermediate $d - 2$ networks have an output dimension of $R \times R$, contributing $O({w^2}l{R^2})$ parameters. Therefore, the total network parameter count is $O({w^2}l(2R + (d - 2){R^2})) = O(d{w^2}l{R^2})$. 
    
\textbf{Tucker-GNO} introduces a learnable core tensor $\mathcal{C}$ to capture complex interactive correlations among variables. Assuming the Tucker ranks across all modes are ${R_i} = R$, the complexity of computing all factor vectors is $O(dw^2l)$, and the multilinear product is realized via step-by-step mode-n contractions, requiring a computational complexity of $O(|\mathcal{C}|)=O(R^d)$. Hence, the single kernel evaluation complexity for Tucker-GNO is $O(dw^2l+R^d)$. Since the output dimensions of the $d$ factor networks are all $R$, the corresponding parameter count is $O(dw^2lR)$. In addition, the parameter count of the core tensor is $O(R^d)$. So the total parameter count for Tucker-GNO is $O(R^d+dw^2lR)$. 

\subsection{Memory Occupancy}
Memory occupancy is a critical consideration in the practical engineering deployment of neural operators. 
In standard GNO, constructing the complete input during integral kernel evaluation consumes $O(\prod_{i=1}^d I_i)$ memory space. 
In the design of FTD-GNO, however, the network input is strictly decoupled into independent variables—$x$, $y$, $a(x)$, and $a(y)$—for decomposed training, and each branch network $\Phi^{(i)}$ only processes the $I_i$ objects within its corresponding dimension. 
This means the model no longer needs to explicitly construct an exponential full-scale tensor of size $\prod_{i=1}^d I_i$; instead, it only needs to independently store $d$ vector representations of length $I_i$. 
Therefore, assuming $I_1=\dots=I_d=I$, the memory footprint during input construction is successfully reduced from the exponential $O(I^d)$ to the linear $O(Id)$. 

Beyond the integral kernel evaluation phase, the spatial integral aggregation phase also faces severe memory consumption challenges. 
Traditional spatial domain integration methods typically rely on edge-based message passing mechanisms, which require explicitly computing and storing the complete integral kernel values on every single edge during execution. 
As grid resolution increases, this edge-level storage requirement easily triggers GPU memory bottlenecks. 
To address this flaw, FTD-GNO ingeniously utilizes the associative property of tensor decomposition to achieve a fundamental reconstruction of the integral formulation. 
The core of this operation lies in isolating all tensor factors that depend exclusively on the central node $x$ (e.g., $A^{(1)}(x)$ and $A^{(3)}(a(x))$). These terms are subsequently factored out of the neighborhood summation $\sum_{y \in N\left( x \right)}$. This algebraic manipulation fundamentally transforms the complex, high-dimensional integrals traditionally computed on graph edges. 
Taking the full-state CP-GNO as an example, its original integral kernel operator takes the following form: 
\begin{align}
    &\left( {\mathcal{K}\left( {a;\Phi } \right){v^{\left( l \right)}}} \right)\left( x \right) \\
    &\qquad\qquad=\frac{1}{|N(x)|} \sum_{y\in N(x)}\left(\sum_{r=1}^{R}A_{r}^{(1)}(x)A_{r}^{(2)}(y)A_{r}^{(3)}(a(x))A_{r}^{(4)}(a(y))\right)v^{(l)}(y),\forall x\in D. \notag 
\end{align}
Through algebraic reconstruction, it is equivalently transformed into 
\begin{align}
    &\left( {\mathcal{K}\left( {a;\Phi } \right){v^{\left( l \right)}}} \right)\left( x \right)\\
    &\qquad\qquad=\frac{1}{|N(x)|}  \sum_{r=1}^{R}A_{r}^{(1)}(x)A_{r}^{(3)}(a(x))\Bigg(\sum_{y_{i}\in N(x)}A_{r}^{(2)}(y)A_{r}^{(4)}(a(y))v^{(l)}(y)\Bigg),\forall x\in D. \notag 
\end{align}

Table \ref{reconstruction} systematically summarizes these algebraic reconstruction formulas across all decomposition formats and input variables. 
This algebraic-level rewriting achieves an effective decoupling of the neighborhood aggregation operation, fundamentally evading the cumbersome process of constructing and storing high-dimensional concatenated tensors along the edge dimension. 
It converts the computation into a highly efficient two-stage process: first, a local aggregation over neighboring nodes $y$, followed by a feature combination at the central node $x$. 
Thanks to this, GPU memory occupancy during network execution transitions from an uncontrollable state highly dependent on the number of edges, smoothly scaling down to a controlled state dependent solely on the number of nodes. 
Notably, we assume a uniform rank $R$ for all tensor factors involved in these decompositions. 

\begin{table}[ht]
\centering
\caption{Algebraic reconstruction formulas of the integral algorithm under CP, TT, and Tucker decomposition formats. $N(x)$ means the neighhourhood of $x$.}
\label{reconstruction} 
\small
\begin{tabular}{ll}
\toprule
Type
& Algebraic Reconstruction Formulation of FTD-GNO \\
\midrule
\multirow{6}{*}{\makecell[c]{CP}}
& $\sum\limits_{r = 1}^R {{A _r^{(1)}}\left( x \right)\bigg( {\sum\limits_{y \in N\left( x \right)} {{A _r^{(2)}}\left( y \right)} } \bigg)}$ \\
& $\sum\limits_{r = 1}^R {{A _r^{(1)}}\left( x \right)\bigg( {\sum\limits_{y \in N\left( x \right)} {{A _r^{(2)}}\left( y \right){v^{\left( l \right)}}\left( y \right)} } \bigg)}$ \\
& $\sum\limits_{r = 1}^R {{A _r^{(1)}}\left( x \right)\bigg( {\sum\limits_{y \in N\left( x \right)} {{A _r^{(2)}}\left( y \right){A _r^{(3)}}\left( {a\left( y \right)} \right)} } \bigg)}$ \\
& $\sum\limits_{r = 1}^R {{A _r^{(1)}}\left( x \right)\bigg(\sum\limits_{y \in N\left( x \right)} {{A _r^{(2)}}\left( y \right){A _r^{(3)}}\left( {a\left( y \right)} \right){v^{\left( l \right)}}\left( y \right)} \bigg)}$ \\
& $\sum\limits_{r = 1}^R {{A _r^{(1)}}\left( x \right){A _r^{(3)}}\left( {a\left( x \right)} \right)\bigg( {\sum\limits_{y \in N\left( x \right)} {{A _r^{(2)}}\left( y \right){A _r^{(4)}}\left( {a\left( y \right)} \right)} } \bigg)}$ \\
& $\sum\limits_{r = 1}^R {{A _r^{(1)}}\left( x \right){A _r^{(3)}}\left( {a\left( x \right)} \right)\bigg(\sum\limits_{y \in N\left( x \right)} {{A _r^{(2)}}\left( y \right){A _r^{(4)}}\left( {a\left( y \right)} \right){v^{\left( l \right)}}\left( y \right)} \bigg)}$ \\
\midrule
\multirow{6}{*}{\makecell[c]{TT}}
& $\sum\limits_{r = 1}^R {{A _r^{(1)}}\left( x \right)\bigg( {\sum\limits_{y \in N\left( x \right)} {{A _r^{(2)}}\left( y \right)} } \bigg)}$ \\
& $\sum\limits_{r = 1}^R {{A _r^{(1)}}\left( x \right)\bigg(\sum\limits_{y \in N\left( x \right)} {{A _r^{(2)}}\left( y \right){v^{\left( l \right)}}\left( y \right)} \bigg)}$ \\
& $\sum\limits_{{r_1},{r_2}} {{A _{{r_1}}^{(1)}}\left( x \right)\left( {\sum\limits_{y \in N\left( x \right)} {{A _{{r_1},{r_2}}^{(2)}}\left( y \right){A _{{r_2}}^{(3)}}\left( {a\left( y \right)} \right)} } \right)}$ \\
& $\sum\limits_{{r_1},{r_2}} {{A _{{r_1}}^{(1)}}\left( x \right)\bigg(\sum\limits_{y \in N\left( x \right)} {{A _{{r_1},{r_2}}^{(2)}}\left( y \right){A _{{r_2}}^{(3)}}\left( {a\left( y \right)} \right){v^{\left( l \right)}}\left( y \right)} \bigg)}$ \\
& $\sum\limits_{{r_1},{r_2},{r_3}} {{A _{{r_1}}^{(1)}}\left( x \right){A _{{r_2},{r_3}}^{(3)}}\left( {a\left( x \right)} \right)\bigg( {\sum\limits_{y \in N\left( x \right)} {{A _{{r_1},{r_2}}^{(2)}}\left( y \right){A _{{r_3}}^{(4)}}\left( {a\left( y \right)} \right)} } \bigg)}$ \\
& $\sum\limits_{{r_1},{r_2},{r_3}} {{A _{{r_1}}^{(1)}}\left( x \right){A _{{r_2},{r_3}}^{(3)}}\left( {a\left( x \right)} \right)\bigg(\sum\limits_{y \in N\left( x \right)} {{A _{{r_1},{r_2}}^{(2)}}\left( y \right){A _{{r_3}}^{(4)}}\left( {a\left( y \right)} \right){v^{\left( l \right)}}\left( y \right)} \bigg)}$ \\
\midrule
\multirow{6}{*}{\makecell[c]{Tucker}}
& $\sum\limits_{{r_1},{r_2}} {{\mathcal{C}_{{r_1},{r_2}}}{A _{{r_1}}^{(1)}}\left( x \right)\bigg( {\sum\limits_{y \in N\left( x \right)} {{A _{{r_2}}^{(2)}}\left( y \right)} } \bigg)}$ \\
& $\sum\limits_{{r_1},{r_2}} {{{\cal C}_{{r_1},{r_2}}}{A _{{r_1}}^{(1)}}\left( x \right)\bigg(\sum\limits_{y \in N\left( x \right)} {{A _{{r_2}}^{(2)}}\left( y \right){v^{\left( l \right)}}\left( y \right)} \bigg)}$ \\
& $\sum\limits_{{r_1},{r_2},{r_3}} {{\mathcal{C}_{{r_1},{r_2},{r_3}}}{A _{{r_1}}^{(1)}}\left( x \right)\bigg( {\sum\limits_{y \in N\left( x \right)} {{A _{{r_2}}^{(2)}}\left( y \right){A _{{r_3}}^{(3)}}\left( {a\left( y \right)} \right)} } \bigg)}$ \\
& $\sum\limits_{{r_1},{r_2},{r_3}} {{{\cal C}_{{r_1},{r_2},{r_3}}}{A _{{r_1}}^{(1)}}\left( x \right)\bigg(\sum\limits_{y \in N\left( x \right)} {{A _{{r_2}}^{(2)}}\left( y \right){A _{{r_3}}^{(3)}}\left( {a\left( y \right)} \right){v^{\left( l \right)}}\left( y \right)} \bigg)}$ \\
& $\sum\limits_{{r_1},{r_2},{r_3},{r_4}} {{\mathcal{C}_{{r_1},{r_2},{r_3},{r_4}}}{A _{{r_1}}^{(1)}}\left( x \right){A _{{r_3}}^{(3)}}\left( {a\left( x \right)} \right)\bigg( {\sum\limits_{y \in N\left( x \right)} {{A _{{r_2}}^{(2)}}\left( y \right){A _{{r_4}}^{(4)}}\left( {a\left( y \right)} \right){v^{\left( l \right)}}\left( y \right)} } \bigg)}$ \\
& $\sum\limits_{{r_1},{r_2},{r_3},{r_4}} {{{\cal C}_{{r_1},{r_2},{r_3},{r_4}}}{A _{{r_1}}^{(1)}}\left( x \right){A _{{r_3}}^{(3)}}\left( {a\left( x \right)} \right)\bigg(\sum\limits_{y \in N\left( x \right)} {{A _{{r_2}}^{(2)}}\left( y \right){A _{{r_4}}^{(4)}}\left( {a\left( y \right)} \right){v^{\left( l \right)}}\left( y \right)} \bigg)}$ \\
\bottomrule
\end{tabular}
\end{table}

\section{Extended Experiments} \label{extended experiments}
\subsection{Experimental Settings} \label{experimental_settings}

This section provides the dataset configurations, model settings, and evaluation protocols supporting the experiments in Section \ref{Sec5}.

\textbf{Datasets.} We consider Darcy flow and two-dimensional Navier–Stokes (NS) on regular grids, together with elasticity on sample-dependent irregular meshes. The Darcy flow and NS benchmark contain 1,000 training samples and 200 test samples. The primary Darcy experiments use resolutions of $16\times16$ and $32\times32$, while the additional architectural comparison is conducted at 16×16. The NS comparison uses resolutions of 16×16 and $128\times128$. For elasticity, each sample is represented by an irregular mesh containing 972 nodes. We use a reduced-data protocol with 200 training samples, 50 validation samples, and the original 200 test samples. Within each controlled comparison, the models share the data split, normalization, and evaluation metric. 

\textbf{Architectures.} The baseline graph operators use four operator layers with 32 hidden channels. For GNO, the internal kernel network $\kappa $ is configured as a three-layer feedforward neural network with hidden layer widths of $[64, 64]$. For the evaluation of FTD-GNO, we utilize CP, TT, and Tucker decompositions for the kernel function. The decomposition rank is fixed at $R=8$ for CP and TT, $R=4$ for Tucker. The hidden layer dimension is set to 64. The rank, width, depth, and kernel-input studies explicitly vary these settings, as specified in the corresponding tables. 

To evaluate extensibility, we construct FNOGNO by replacing the preceding graph operator layers with Fourier layers while retaining the final graph integral. CP-, TT-, and Tucker-FNOGNO replace only the kernel of this final layer. Each FNOGNO/FTD-FNOGNO comparison therefore shares the Fourier backbone, lifting and projection settings, and final graph neighborhood. FNO is included as an architectural reference. 

To distinguish functional factorization from weight compression, we additionally consider LR-GNO. Each dense weight matrix $W_l\in\mathbb{R}^{m_l\times n_l}$ in its kernel MLP is parameterized as
\begin{align}
    W_l = A_l B_l,\quad A_l\in\mathbb{R}^{m_l\times r_l},\quad B_l\in\mathbb{R}^{r_l\times n_l}.
\end{align}

The integer ranks $r_l$ are selected to approximately match the parameter count of FTD-GNO. LR-GNO retains the same graph-integral architecture and edge-wise kernel evaluation as GNO. 

\textbf{Optimization and training budgets.} In all experiments, we train the models using the AdamW optimizer. The batch size is set to 32, with an initial learning rate of $1e - 3$ and a weight decay of $1e - 4$. We also employ a cosine annealing schedule for learning rate decay.  The primary Darcy baseline and resolution-transfer studies, together with the parameter matched comparison in Table \ref{Tab:LR-GNO}, use 300 training epochs. The additional Darcy/NS architectural comparisons in Table \ref{Tab:more baseline} and the capacity studies in Tables \ref{Tab:rank1} and \ref{Tab:rank2} use 100 epochs. The elasticity study reports all models at 100 epochs and additional runs of the factorized variants at 300 epochs. The extended Darcy study varies the CP-GNO training budget while keeping its architecture fixed. These experiments are analyzed separately to distinguish equal-epoch comparisons from additional optimization. 

\textbf{Error metric.} The ${l_2}$ relative error is utilised in both the training optimization and evaluation. 
Given a dataset of input-output pairs $\left\{ {{a_i},{u_i}} \right\}_{i = 1}^N$ and a model ${\mathcal{G}_\theta }:\mathcal{A} \to \mathcal{U}$, the relative error is calculated as outlined below: 
\begin{equation}
    L\left( {{\mathcal{G}_\theta }\left( a \right),u} \right) = \frac{1}{N}\sum\limits_{i = 1}^N {\frac{{{{\left\| {{\mathcal{G}_\theta }\left( {{a_i}} \right) - {u_i}} \right\|}^2}}}{{{{\left\| {{u_i}} \right\|}^2}}}}. 
\end{equation}

\textbf{Resource measurements.} Experiments are conducted on a single NVIDIA GeForce RTX 5090 GPU using PyTorch 2.1.0. Total training time is measured around the complete training call, with CUDA synchronization before and after timing. It includes minibatch loading, preprocessing performed during training, forward and backward computation, in-model neighborhood operations, optimization, and scheduled validation. Initial dataset download and model construction are excluded. Peak GPU memory is reported in MB, and OOM denotes a configuration that exceeds the available memory under the tested setup. Point estimates and repeated run mean ± standard deviation measurements are presented separately. 

\subsection{Additional Experimental Results} \label{additional experimental}

\subsubsection{Detailed analysis in Section \ref{Baseline Comparison}} \label{sec:E.2.1}

\textbf{Weight compression.} Table \ref{Tab:LR-GNO} compares GNO, parameter matched LR-GNO, and the three FTD-GNO variants on Darcy flow. LR-GNO and CP/TT-GNO have almost identical parameter counts, but their peak-memory costs differ substantially. LR-GNO compresses the kernel MLP weights while continuing to evaluate the network on neighboring node pairs. In contrast, FTD reorganizes kernel evaluation and aggregation through reusable node-wise factors. This comparison indicates that reducing the number of parameters alone does not explain the memory advantage.

The prediction errors of LR-GNO and CP/TT-GNO are close, although LR-GNO performs slightly better. Thus, functional factorization introduces an accuracy–resource trade-off even at matched parameter counts. The result supports the contribution of the factorized contraction without implying identical approximation or optimization behavior between the two parameterizations. 

\begin{table*}[ht]
    \caption{Parameter matched comparison on Darcy flow at $16\times16$ and 300 epochs. Peak memory is reported in MB. }
    \label{Tab:LR-GNO}
    \centering 
    \small
    \begin{tabular}{l c c c}
        \toprule
        Model & Parameters & Error & Peak memory \\
        \midrule
        GNO & 95265 & 0.1872 & 420.6 \\
        LR-GNO & 74785 & 0.1946 & 479.8 \\
        CP-GNO & 74725 & 0.1975 & 182.7 \\
        TT-GNO & 74725 & 0.1975 & 182.7 \\
        Tucker-GNO & 72709 & 0.2157 & 183.7 \\
        \bottomrule
    \end{tabular}
\end{table*}

\textbf{Regular grid benchmarks.} Table \ref{Tab:more baseline} reports the complete architectural comparison on Darcy flow and NS under the 100 epochs protocol. Across the two datasets, FTD reduces peak memory relative to the corresponding unfactorized baselines in both the graph-only and Fourier–graph groups. The three FTD-FNOGNO variants maintain accuracy close to FNOGNO in the low-resolution comparisons. These results show that FTD remains effective when graph integration is combined with Fourier operator layers. 

At the higher NS resolution, GNO and FNOGNO encounter OOM, whereas their factorized variants remain trainable. This extends the feasible resolution of the tested graph-integral architectures. Nevertheless, the graph-only configurations exhibit relatively high NS errors, and FNO provides more accurate predictions with lower memory consumption than the tested hybrids. Incorporating Fourier layers improves the predictive performance of the graph-based configurations, but does not eliminate their gap relative to FNO. The experiments therefore support FTD as an efficient graph-integral component whose final accuracy depends on the surrounding architecture and PDE task. 

\textbf{Irregular meshes.} Table \ref{Tab:irregular} evaluates FTD-GNO on the elasticity benchmark with sample-dependent irregular meshes. In accordance with the 100-epoch protocol, all three decomposition variants substantially reduce peak memory and recorded total training time relative to GNO. The additional 300-epoch runs improve the factorized models beyond the error of the 100-epoch GNO while retaining lower memory and recorded total runtime. The findings indicate that the reduced training cost can be used to support additional optimization. Together, these results suggest that the efficiency benefit applies to irregular discretizations as well as regular grids.

\begin{table*}[ht]
    \caption{Elasticity on sample-dependent irregular meshes with 972 nodes. }
    \label{Tab:irregular}
    \centering 
    \small
    \begin{tabular}{l c c c c c}
        \toprule
        Model & Discretization & Epochs & Error & Peak memory & Time (s) \\
        \midrule
        GNO        & 972 irregular nodes & 100 & 0.2800 & 263.79 & 17640.04 \\
        CP-GNO     & 972 irregular nodes & 100 & 0.3021 & 61.74 & 517.41 \\
        TT-GNO     & 972 irregular nodes & 100 & 0.3021 & 61.74 & 521.04 \\
        Tucker-GNO & 972 irregular nodes & 100 & 0.3154 & 63.23 & 502.32 \\
        \midrule
        CP-GNO  & 972 irregular nodes & 300 & 0.2623 & 61.74 & 1542.86 \\
        TT-GNO  & 972 irregular nodes & 300 & 0.2623 & 61.74 & 1513.48 \\
        Tucker-GNO  & 972 irregular nodes & 300 & 0.2709 & 63.23 & 1557.15 \\
        \bottomrule
    \end{tabular}
\end{table*}

\begin{table*}[ht]
    \caption{Extended training study on Darcy flow at $16\times16$. }
    \label{Tab:epoch}
    \centering 
    \small
    \begin{tabular}{l c c c}
        \toprule
        Model & Epochs & Error & Time (s) \\
        \midrule
        GNO & 300 & 0.1872 & 6542.28 \\
        CP-GNO & 300 & 0.1975 & 647.00 \\
        CP-GNO & 500 & 0.1971 & 1081.49 \\
        CP-GNO & 700 & 0.1931 & 1603.67 \\
        CP-GNO & 1000 & 0.1926 & 2120.37 \\
        CP-GNO & 2000 & 0.1902 & 4521.40 \\
        CP-GNO & 3000 & 0.1877 & 6423.93 \\
        \bottomrule
    \end{tabular}
\end{table*}

\subsubsection{Supplementary experiments} \label{sec:E.2.2}

\textbf{Capacity: Whether longer training or greater model capacity improves accuracy.} 

For Darcy flow, longer training progressively narrows the accuracy gap between CP-GNO and GNO while leaving the CP-GNO architecture unchanged. At the longest tested budget, CP-GNO approaches the baseline accuracy, but a small error gap remains and their recorded total training times become comparable, shown in Table \ref{Tab:epoch}. The elasticity results in Table \ref{Tab:irregular} likewise demonstrate the value of additional optimization. 

Table \ref{Tab:rank1} examines factor network width $w$, depth $l$, and decomposition rank $R$ under a fixed 100-epoch budget. Increasing width provides little accuracy improvement in the tested configurations while increasing parameter count and memory. At the tested width and rank, the shallowest configuration achieves both lower error and lower measured resource consumption than the deeper alternatives. Increasing rank also produces a non-monotonic error response. The study in Table \ref{Tab:rank2} further shows that the preferred rank depends on the decomposition. CP and TT achieve their lowest tested error at $R=8$, whereas Tucker performs best at $R=4$. 

\begin{table*}[ht]
    \caption{CP-GNO sensitivity to factor-network width, depth and rank on Darcy flow. }
    \label{Tab:rank1}
    \centering 
    \small
    \begin{tabular}{c c c c c c c c}
        \toprule
        Ablation & $w$ & $l$ & $R$ & Error & Parameters & Peak memory & Time (s) \\
        \midrule
        $w$ & 16 & 2 & 8 & 0.2269 & 24293 & 152.82 & 205.09 \\
        $w$ & 32 & 2 & 8 & 0.2250 & 48101 & 189.09 & 214.35 \\
        $w$ & 64 & 2 & 8 & 0.2259 & 108005 & 269.38 & 259.47 \\
        $l$ & 64 & 1 & 8 & 0.2114 & 74725 & 197.02 & 232.84 \\
        $l$ & 64 & 2 & 8 & 0.2259 & 108005 & 269.38 & 259.47 \\
        $l$ & 64 & 3 & 8 & 0.2308 & 141285 & 334.16 & 283.20 \\
        $R$ & 64 & 1 & 4 & 0.2171 & 72645 & 201.77 & 233.54 \\
        $R$ & 64 & 1 & 8 & 0.2114 & 74725 & 197.02 & 232.84 \\
        $R$ & 64 & 1 & 16 & 0.2276 & 78885 & 221.16 & 239.54 \\
        \bottomrule
    \end{tabular}
\end{table*}
\begin{table*}[ht]
    \caption{Rank sensitivity of different decomposition methods. }
    \label{Tab:rank2}
    \centering 
    \small
    \begin{tabular}{c c c c}
        \toprule
        $R$ & CP $L_2$ error & TT $L_2$ error & Tucker $L_2$ error \\
        \midrule
        2 & 0.2161 & 0.2161 & 0.2177 \\
        4 & 0.2171 & 0.2171 & 0.2104 \\
        8 & 0.2114 & 0.2114 & 0.2257 \\
        16 & 0.2276 & 0.2276 & 0.2134 \\
        \bottomrule
    \end{tabular}
\end{table*}

These observations indicate that additional optimization can improve the accuracy of a fixed-capacity factorized model, while increasing capacity alone does not guarantee better finite-budget performance. In practice, rank, width, depth, and training budget should be selected jointly using validation performance and the available memory and runtime budget. 

\textbf{Kernel inputs: Whether different kernel inputs affect memory costs.} 

Beyond spatial grid topology, the chosen integration type, which determines the input variable dimensions, fundamentally dictates the complexity of multivariate coupling and the overall memory envelope. Table \ref{Tab5:peak_memory2} examines memory consumption of different kernel inputs, with resolution, radius, and rank held fixed. When evaluated on basic two spatial inputs $\kappa(x,y)$, a rank-$R$ TT decomposition with boundary ranks $R_0=R_2=1$ satisfies 
\begin{align}
    \kappa_{\mathrm{TT}}(x,y)=\sum_{r=1}^{R}A_{1,r}^{(1)}(x)A_{r,1}^{(2)}(y),
\end{align}
which is exactly the rank-$R$ CP form 
\begin{align}
    \kappa_{\mathrm{CP}}(x,y)=\sum_{r=1}^{R}A_r^{(1)}(x)A_r^{(2)}(y).
\end{align}
Therefore, when the factor subnetworks are configured identically, CP and TT represent the same function class and exhibit identical memory footprints. As the number of nuclear inputs increases, their costs begin to differ. When the input expands to encompass the full state $(x,y,a(x),a(y))$, CP-GNO and TT-GNO demonstrate exceptional robustness, maintaining highly compressed memory footprints with only marginal increases. In contrast, the memory consumption of Tucker-GNO surges significantly to 861.5 MB, an inflation intrinsically linked to the combinatorial storage requirements of its requisite core tensor. Consequently, for operator learning tasks involving complex multivariable inputs, CP and TT decompositions emerge as the superior architectural paradigms, delivering an optimal balance of parameter efficiency and memory compression. 

\begin{table*}[ht]
    \centering
    \small
    \caption{Peak GPU memory footprint (MB) versus the number of input variables, tested at grid resolution $16 \times 16$, radius $r = 0.08$, and decomposition rank $R = 8$.}
    \label{Tab5:peak_memory2}
    \begin{tabular}{lccc}
        \toprule
        Model & $(x,y)$ & $(x,y,a(y))$ & $(x,y,a(x),a(y))$ \\
        \midrule
        GNO         & 420.6 & 432.1 & 426.7 \\
        CP-GNO      & 182.7 & 203.1 & 223.2 \\
        TT-GNO      & 182.7 & 217.3 & 251.6 \\
        Tucker-GNO  & 183.7 & 211.2 & 861.5 \\
        \bottomrule
    \end{tabular}
\end{table*}

\textbf{Statistical Significance and Stability.} 

To rigorously validate the hardware stability of our proposed algebraic reconstruction, we report detailed peak GPU memory footprints across varying grid resolutions and radii in Table \ref{tab:mean_peak_memory}. To ensure statistical significance, all measurements are reported as mean $\pm$ standard deviation derived from multiple independent runs with different random seeds.The results reveal that FTD-GNO variants not only achieve substantial memory reductions compared to the standard GNO but also exhibit exceptionally low variance, with standard deviations generally remaining below 1.0 MB. 
\begin{table*}[ht]
\centering
\caption{Peak GPU memory footprint (MB) of different models under varying resolutions and radii. Results are consistently reported as mean $\pm$ standard deviation across independent runs to ensure statistical significance. }
\label{tab:mean_peak_memory}
\setlength{\tabcolsep}{4pt}
\small
\resizebox{\textwidth}{!}{
\begin{tabular}{lccccccccc}
\toprule
& \multicolumn{3}{c}{$16 \times 16$}
& \multicolumn{3}{c}{$32 \times 32$}
& \multicolumn{3}{c}{$64 \times 64$} \\
\cmidrule(lr){2-4} \cmidrule(lr){5-7} \cmidrule(lr){8-10}
Model
& $r{=}0.08$ & $r{=}0.12$ & $r{=}0.16$
& $r{=}0.08$ & $r{=}0.12$ & $r{=}0.16$
& $r{=}0.08$ & $r{=}0.12$ & $r{=}0.16$ \\
\midrule
GNO
& \makecell[c]{421.13  \\$\pm$0.68} & \makecell[c]{696.80   \\$\pm$0.62} & \makecell[c]{1511.53  \\$\pm$0.68}
& \makecell[c]{6427.10 \\$\pm$0.62} & \makecell[c]{13209.40 \\$\pm$0.62} & \makecell[c]{19733.80 \\$\pm$0.62}
& \textsc{OOM}                      & \textsc{OOM}                       & \textsc{OOM} \\

CP-GNO ($R{=}8$)
& \makecell[c]{183.20  \\$\pm$0.62} & \makecell[c]{183.23   \\$\pm$0.68} & \makecell[c]{183.43   \\$\pm$0.68}
& \makecell[c]{680.87  \\$\pm$0.67} & \makecell[c]{682.50   \\$\pm$0.62} & \makecell[c]{684.10   \\$\pm$0.62}
& \makecell[c]{2689.77 \\$\pm$0.95} & \makecell[c]{2715.33  \\$\pm$0.67} & \makecell[c]{2753.67  \\$\pm$0.47} \\
CP-GNO ($R{=}16$)
& \makecell[c]{275.20  \\$\pm$0.62} & \makecell[c]{275.30   \\$\pm$0.62} & \makecell[c]{275.50   \\$\pm$0.62}
& \makecell[c]{1048.90 \\$\pm$0.62} & \makecell[c]{1050.53  \\$\pm$0.68} & \makecell[c]{1052.13  \\$\pm$0.68}
& \makecell[c]{4161.77 \\$\pm$0.95} & \makecell[c]{4187.80  \\$\pm$0.61} & \makecell[c]{4225.70  \\$\pm$0.53} \\
CP-GNO ($R{=}32$)
& \makecell[c]{459.30  \\$\pm$0.62} & \makecell[c]{459.40   \\$\pm$0.62} & \makecell[c]{459.60   \\$\pm$0.62}
& \makecell[c]{1785.00 \\$\pm$0.62} & \makecell[c]{1786.63  \\$\pm$0.68} & \makecell[c]{1788.20  \\$\pm$0.62}
& \makecell[c]{7104.93 \\$\pm$0.45} & \makecell[c]{7132.07  \\$\pm$0.50} & \makecell[c]{7169.57  \\$\pm$0.76} \\

TT-GNO ($R{=}8$)
& \makecell[c]{183.20  \\$\pm$0.62} & \makecell[c]{183.23   \\$\pm$0.68} & \makecell[c]{183.43   \\$\pm$0.68}
& \makecell[c]{680.87  \\$\pm$0.67} & \makecell[c]{682.50   \\$\pm$0.62} & \makecell[c]{684.10   \\$\pm$0.62}
& \makecell[c]{2689.77 \\$\pm$0.95} & \makecell[c]{2715.33  \\$\pm$0.67} & \makecell[c]{2753.67  \\$\pm$0.47} \\
TT-GNO ($R{=}16$)
& \makecell[c]{275.20  \\$\pm$0.62} & \makecell[c]{275.30   \\$\pm$0.62} & \makecell[c]{275.50   \\$\pm$0.62}
& \makecell[c]{1048.90 \\$\pm$0.62} & \makecell[c]{1050.53  \\$\pm$0.68} & \makecell[c]{1052.13  \\$\pm$0.68}
& \makecell[c]{4161.77 \\$\pm$0.95} & \makecell[c]{4187.80  \\$\pm$0.61} & \makecell[c]{4225.70  \\$\pm$0.53} \\
TT-GNO ($R{=}32$)
& \makecell[c]{459.30  \\$\pm$0.62} & \makecell[c]{459.40   \\$\pm$0.62} & \makecell[c]{459.60   \\$\pm$0.62}
& \makecell[c]{1785.00 \\$\pm$0.62} & \makecell[c]{1786.63  \\$\pm$0.68} & \makecell[c]{1788.20  \\$\pm$0.62}
& \makecell[c]{7104.93 \\$\pm$0.45} & \makecell[c]{7132.07  \\$\pm$0.50} & \makecell[c]{7169.57  \\$\pm$0.76} \\

Tucker-GNO ($R{=}8$)
& \makecell[c]{184.20  \\$\pm$0.62} & \makecell[c]{184.23   \\$\pm$0.68} & \makecell[c]{184.43   \\$\pm$0.68}
& \makecell[c]{684.90  \\$\pm$0.62} & \makecell[c]{686.50   \\$\pm$0.62} & \makecell[c]{688.10   \\$\pm$0.62}
& \makecell[c]{2706.43 \\$\pm$0.55} & \makecell[c]{2731.03  \\$\pm$0.90} & \makecell[c]{2769.80  \\$\pm$0.46} \\
Tucker-GNO ($R{=}16$)
& \makecell[c]{277.23  \\$\pm$0.68} & \makecell[c]{277.30   \\$\pm$0.62} & \makecell[c]{277.50   \\$\pm$0.62}
& \makecell[c]{1056.93 \\$\pm$0.68} & \makecell[c]{1058.57  \\$\pm$0.67} & \makecell[c]{1060.13  \\$\pm$0.68}
& \makecell[c]{4193.80 \\$\pm$0.90} & \makecell[c]{4219.30  \\$\pm$0.62} & \makecell[c]{4257.87  \\$\pm$0.55} \\
Tucker-GNO ($R{=}32$)
& \makecell[c]{463.37  \\$\pm$0.67} & \makecell[c]{463.43   \\$\pm$0.68} & \makecell[c]{463.63   \\$\pm$0.68}
& \makecell[c]{1801.07 \\$\pm$0.67} & \makecell[c]{1802.70  \\$\pm$0.62} & \makecell[c]{1804.27  \\$\pm$0.67}
& \makecell[c]{7168.97 \\$\pm$0.50} & \makecell[c]{7196.13  \\$\pm$0.45} & \makecell[c]{7233.63  \\$\pm$0.71} \\
\bottomrule
\end{tabular}}
\end{table*}

\section{Limitations} \label{limitations}

Our limitations and future work are summarized as follows:

(1) \textbf{High-dimensional scalability of the Tucker architecture:} When handling inputs with more than four variables, the core tensor in Tucker-GNO remains susceptible to a combinatorial explosion in memory footprint. 
While CP and TT decompositions successfully avoid this issue, optimizing the Tucker architecture to fully harness its powerful cross-dimensional coupling capabilities requires further investigation. 

(2) \textbf{Representation bottlenecks of static ranks:} Currently, the tensor decomposition rank $R$ must be pre-defined as a hyperparameter. 
Empirical evidence shows that $R=8$ is sufficient for most tasks. However, when modeling highly dynamic and complex physical processes (e.g., strong turbulent fields), a fixed rank may constrain the expressive capacity of the integral kernel. 

(3) \textbf{Residual memory risks under dense topologies:} Despite deep decomposition and decoupling at the input variable level, FTD-GNO may still face potential OOM risks if the number of neighboring nodes increases drastically. 

To address these limitations, future research will explore Dynamic Rank Allocation mechanisms, with a view to enabling the model to adaptively accommodate the complexities of varying physical fields. 
Additionally, we will investigate spatial decomposition models based directly on grid coordinates. 
The ultimate goal is to fundamentally eradicate the OOM risks inherent to GNOs when processing fully connected graphs or exceptionally large receptive fields. 

\end{document}